\documentclass{article}
\usepackage{amsfonts}
\usepackage{graphicx} 
\usepackage[margin=1in]{geometry}
\usepackage{amsmath,amsthm,amsfonts,amssymb,mathtools}
\usepackage[linesnumbered,ruled,vlined]{algorithm2e}
\usepackage[colorlinks,linkcolor = blue]{hyperref}
\usepackage[capitalise]{cleveref}
\usepackage{tikz}
\usepackage{booktabs}
\usepackage{float}
\usetikzlibrary{positioning,arrows,calc}

\DeclareMathOperator{\EE}{\mathbb{E}}

\newcommand{\CoinBiasEstimationProblem}{\allowbreak\textup{\textsc{CoinBiasEstimationProblem}}}
\DeclareMathOperator{\poly}{\textup{poly}}

\DeclareMathOperator*{\argmax}{arg\,max}

\DeclareMathOperator{\Unif}{\textup{Unif}}
\DeclareMathOperator{\Bern}{\textup{Bern}}

\DeclareRobustCommand{\dTV}{\textup{d}_{\textup{TV}}}

\DeclareMathOperator{\DKL}{{\textup{D}}_{\textup{KL}}}

\newcommand{\boldp}{{\mathbf{p}}}
\newcommand{\boldq}{{\mathbf{q}}}

\newcommand{\boldx}{{\mathbf{x}}}
\newcommand{\boldy}{{\mathbf{y}}}

\newcommand{\calA}{{\mathcal{A}}}
\newcommand{\calB}{{\mathcal{B}}}

\newcommand{\calD}{{\mathcal{D}}}

\newcommand{\calL}{{\mathcal{L}}}

\newcommand{\calP}{{\mathcal{P}}}

\newcommand{\calR}{{\mathcal{R}}}

\newcommand{\calX}{{\mathcal{X}}}
\newcommand{\calY}{{\mathcal{Y}}}

\providecommand{\Description}[1]{}
\usepackage{pgfplots}
\pgfplotsset{compat=1.7}

\theoremstyle{definition}
\newtheorem{definition}{Definition}[section]

\theoremstyle{plain}
\newtheorem{theorem}[definition]{Theorem}

\newtheorem{lemma}[definition]{Lemma}

\newcommand{\email}[1]{\href{mailto:#1}{\texttt{#1}}}

\title{The Price of List Replicability: Tight Bounds for Bias Estimation}
\author{
    Michael Chen\\
    Iowa State University\\
    \email{mqychen@iastate.edu}
    \and
    Atieh Bastan\\
    Iowa State University\\
    \email{atie@iastate.edu}
    \and
    A. Pavan\\
    Iowa State University\\
    \email{pavan@iastate.edu}
    \and
    N. V. Vinodchandran\\
    University of Nebraska-Lincoln\\
    \email{vinod@unl.edu}
}

\date{November 2, 2026}
\begin{document}
\maketitle
\begin{abstract}
We study the sample complexity of list replicable algorithms for estimating the biases of $d$ coins. An algorithm is $(k,\varepsilon,\delta)$-list replicable if for every bias vector $\mathbf{p}\in[0,1]^d$, there is a set of outputs $L_{\mathbf{p}}\subseteq \mathbb{R}^d$ of at most $k$ elements such that every $\mathbf{y}\in L_{\mathbf{p}}$ satisfies $\lVert\mathbf{y}-\mathbf{p}\rVert_\infty\leq\varepsilon$, and the algorithm output is in $L_{\mathbf{p}}$ with probability at least $1-\delta$. Prior work gave a $(d+1,\varepsilon,\delta)$-list replicable algorithm with sample complexity $\widetilde{O}\left(\frac{d^3}{\varepsilon^2}\right)$, but no nontrivial sample complexity lower bound was known.

We establish the first non-trivial sample complexity lower bound for this problem and give an improved algorithm that matches it up to polylogarithmic factors. Together, these results determine the optimal sample complexity for list size $d+1$ to be $\widetilde{\Theta}\left(\frac{d^2}{\varepsilon^2}\right)$, showing that the quadratic dependence on dimension is inherent up to polylogarithmic factors. More generally, for every $k\in[d+1,2^d]$, we establish matching upper and lower bounds of $\widetilde{\Theta}\left(\frac{d^2}{\varepsilon^2\log k}\right)$ on the total sample complexity. 
\end{abstract}

\section{Introduction}
Typically, learning algorithms observe samples from an unknown distribution and produce an output. In general, such algorithms are not {\em replicable:} two different runs of the algorithm observe two different samples (though drawn from the same distribution), resulting in potentially different outputs. Motivated by this, various notions of replicable learning have been introduced in the recent literature. In particular, the work of ~\cite{ILPS2022} formalized a notion of replicability called {\em ${\rho}$-replicability}, and the works of \cite{DPVV2023} and \cite{CMY2023} introduced the notion of {\em list replicability}. 
Since then, the notion of replicable learning has received considerable attention and has been studied in various contexts such as PAC learning, Reinforcement learning, multi-armed bandits, etc. \cite{HIKLC2024,BGH+2023,BLM2020,EKK+2023,DGSB2026,EHKS2023,HM2025,DPVV2022,BHHLT2026}. A basic problem studied in this area is $d$-$\CoinBiasEstimationProblem$: given sample access to $d$ coins with unknown biases, estimate every bias to within additive error $\varepsilon$. The $d$-coin bias estimation problem is a basic instance of high-dimensional mean estimation over Bernoulli product distributions and captures central challenges in designing replicable algorithms and understanding their limitations.

In this work, we focus on \textit{list replicable} \cite{DPVV2023, CMY2023} algorithms for $d$-$\CoinBiasEstimationProblem$. Informally, an algorithm is $k$-list replicable if, given sample access to an unknown distribution $\calD$, it outputs with high probability an element of a set $L_\calD$ containing at most $k$ correct answers. This captures replicability by restricting outputs to a small set that depends only on $\calD$, not on the observed samples; smaller $k$ gives a stronger guarantee.
Ideally, we want the list size to be $k=1$. However, as observed in \cite{DPVV2023}, such a strong notion of replicability is impossible even for the case of single coin bias estimation. 

For the $d$-$\CoinBiasEstimationProblem$, it is known that there is no $d$-list replicable algorithm, but there is a $(d+1)$-list replicable algorithm~\cite{DPVV2023}. The $(d+1)$-list replicable algorithm obtains a $\frac{\epsilon}{2d}$-close estimate of each coin and rounds the resulting vector using a partition scheme known as secluded partitions. Thus, the sample complexity of the $(d+1)$-list replicable algorithm is $\widetilde{O}\left(\frac{d^3}{\varepsilon^2}\right)$. Note that if we do not require replicability, then there is a straightforward algorithm whose sample complexity is only $\widetilde{O}\left(\frac{d}{\varepsilon^2}\right)$: toss each coin $\widetilde{\Theta}\left(\frac{1}{\varepsilon^2}\right)$ times to obtain an $\varepsilon$-close estimate of its bias. This raises the question of whether there is a {\em price to achieve replicability}. {\em Do replicable algorithms require inherently higher sample complexity?} 
Motivated by this, we ask the following question:
\begin{center}
\textit{What is the optimal sample complexity of estimating the biases of $d$ coins with $\ell_\infty$ error at most $\varepsilon$ in a $(d+1)$-list replicable manner?}
\end{center}

We resolve this question up to polylogarithmic factors: we show that the sample complexity of list replicable algorithms for $d$-$\CoinBiasEstimationProblem$ is $\widetilde{\Theta}\left(\frac{d^2}{\varepsilon^2}\right)$. In particular, our upper bound improves the polynomial dependence on $d$ from cubic to quadratic \cite{DPVV2023}, and our lower bound result shows that this improved upper bound is optimal up to polylogarithmic factors. {\em Thus, list replicability comes at a price: a factor-\(d\) increase in sample complexity, up to polylogarithmic factors.}

We generalize this result to arbitrary $k \in [d+1, 2^d]$, by showing that the sample complexity of $k$-list replicable algorithms for the problem is $\widetilde{\Theta}\left(\frac{d^2}{\varepsilon^2\log k}\right)$.

\section{Related Works}

\paragraph{List replicability.}
List replicability was introduced independently in \cite{DPVV2023,CMY2023}, in the contexts of coin bias estimation and PAC learning. The works most closely related to ours are \cite{DPVV2023,WDPRV2024}, which study list replicable estimation through geometric partitions and establish bounds on list size and on the sample complexity of the resulting algorithms. These works give a $(d+1,\varepsilon,\delta)$-list replicable algorithm for $d$-$\CoinBiasEstimationProblem$ with sample complexity $\widetilde{O}\left(\frac{d^3}{\varepsilon^2}\right)$, but leave open the optimal sample complexity. Our work establishes matching upper and lower bounds up to polylogarithmic factors and characterizes the tradeoff between list size and sample complexity. Subsequent work has studied list replicability in other contexts, including for large-margin halfspaces~\cite{BHHLT2026}, multi-armed bandits~\cite{CPVWY2025}, and reinforcement learning~\cite{ZCPVYW2025}.

\paragraph{$\rho$-replicability.} The notion of $\rho$-replicability was introduced in \cite{ILPS2022}. Here, a learning algorithm observes samples from a distribution and has internal randomness.  An algorithm is $\rho$-replicable if two executions of the algorithm on independent samples from the same distribution, using the same internal random seed, produce identical outputs with probability at least $1-\rho$. The probability is over both the samples and the shared seed. This notion has been studied in learning and statistical estimation~\cite{ILPS2022,BGH+2023,HIKLC2024}, multi-armed bandits~\cite{EKK+2023,DGSB2026}, and reinforcement learning~\cite{EHKS2023}. Of particular relevance, \cite{HIKLC2024} establishes nearly matching sample complexity bounds for high-dimensional mean estimation under $\ell_2$ and $\ell_\infty$ error. For $d$-$\CoinBiasEstimationProblem$, these results imply a lower bound of $\widetilde{\Omega}\left(\frac{d^2}{\varepsilon^2\rho^2}\right)$ total coin tosses for nonadaptive algorithms, with a matching upper bound up to polylogarithmic factors in the relevant parameter regime. While $\rho$-replicability requires agreement between runs using shared randomness, list replicability requires concentration on a small set of outputs determined solely by the underlying distribution.

\paragraph{Stability, privacy, and randomness.} Global stability, introduced in \cite{BLM2020} in the study of private learning, requires that, for each distribution, some fixed accurate hypothesis be output with probability bounded below. An equivalence between global stability and list replicability in PAC learning was established in \cite{CMY2023}. For a broad class of statistical problems, \cite{BGH+2023} gives sample-efficient reductions between shared-randomness replicability, approximate differential privacy, and perfect generalization, while also establishing computational separations between these notions. Certificate replicability, introduced
in \cite{DPVV2023}, provides another way to quantify the information needed to reproduce an output. The randomness complexity of replicability and differential privacy, and its relationship to global stability, were further studied in \cite{HM2025}.

\paragraph{Pseudodeterminism and reproducible computation.} The study of replicability in learning has its roots in pseudodeterminism in the context of randomized algorithms. Pseudodeterminism was introduced in \cite{GG2011} and further developed in \cite{GGR2013}. A randomized algorithm $\calA$ is pseudodeterministic if for every input  $x$, there is a canonical value $c_x$ such that $\calA$ on input $x$ outputs $c_x$ with high probability.  Multi-pseudodeterminism~\cite{Goldreich2025} relaxes this requirement by allowing a small set of canonical answers: for every $x$, there is a canonical list $L_x$ such that the output of $\calA$ on input $x$ belongs to $L_x$ with high probability. If the list $L_x$ is of size at most $k$, then $\calA$ is a $k$-pseudodeterministic algorithm. The notion of list replicability is analogous to multi-pseudodeterminism in the context of learning: the canonical set depends on the unknown distribution, and the guarantee must hold across independently drawn samples. A related notion is that of algorithms with few influential random bits~\cite{GL2019}. Let $\calA$ be a randomized algorithm that uses $m$ random bits. Separate the randomness into two parts $rs$, where $|r| = \ell$ and $|s| = m-\ell$. The algorithm $\calA$ is $\ell$-bit influential if for every input $x$, and for most choices of $\ell$-bit strings $r$, there is a canonical value $c_{x,r}$ such that $\calA(x, rs) = c_{x,r}$ occurs with high probability when $s$ is randomly chosen. The notion of pseudodeterminism has been investigated in streaming algorithms~\cite{GGMW2020,BKKS2023,GGS2023,CCPV2025}, communication complexity~\cite{BHHLT2026,GHR+2026}, and complexity theory, including connections to probabilistic time, promise problems, and pseudodeterministic constructions~\cite{LOS2021,DPVV2022,CLORS2026}.

\section{Preliminaries}
\subsection{Notation}
We use bold lowercase letters to denote vectors $\boldp,\boldq,\boldx,\boldy$. We index them using their non-bold counterpart, e.g., $\boldp=(p_1,\ldots,p_d)$. We say $\hat \boldp$ is $\varepsilon$-close to $\boldp$ to denote the condition $\lVert\hat\boldp -\boldp\rVert_\infty\leq\varepsilon$.  The condition $\boldp\leq \boldq$ means pointwise $\leq$, i.e., $\forall i\ldotp p_i\leq q_i$. And when we say $\boldp\leq\mathbf{\frac{3}{4}}$ we mean $\forall i\ldotp p_i\leq \frac{3}{4}$. For a vector $\boldp\in\mathbb{R}^d$ and radius $r>0$, we use $\calB_{\infty}^{\circ}(r,\boldp)$ to denote an open ball of radius $r$ with respect to $\ell_\infty$ distance, i.e., $\calB_{\infty}^{\circ}(r,\boldp)=\{\boldq\in\mathbb{R}^d:\lVert \boldp-\boldq\rVert_\infty<r\}$. For $n\in\mathbb{N}$ we let $[n]=\{1,\ldots, n\}$ and for $a,b\in\mathbb{R}$ we use $[a,b]$, $(a,b)$, $[a,b)$, and $(a,b]$ to represent the closed, open intervals, and their corresponding half-open intervals, respectively. We use $\widetilde{O}(\cdot)$, $\widetilde{\Omega}(\cdot)$, $\widetilde\Theta(\cdot)$ to subsume polylogarithmic factors, while keeping the dependence on the list size $k$ explicit.


For distributions $P,Q$ on the same measurable space, we use $\frac{dQ}{dP}$ to denote the Radon-Nikodym derivative; e.g., if $P, Q$ are discrete distributions on $\calX$ then for all $x\in\calX$, it holds that $\frac{dQ}{dP}(x)=\frac{Q(x)}{P(x)}$, i.e., it is simply the ratio of their probability mass functions. We use $Q\ll P$ to denote the condition that for all measurable sets $A$, if $P(A)=0$ then $Q(A)=0$.

\subsection{Sampling Algorithms}
We work in the sampling model and consider randomized algorithms that have sample access to $d\in\mathbb{N}$ coins with unknown bias vector $\boldp=(p_1,\ldots,p_d)\in[0,1]^d$. At each stage, an algorithm $\calA$ may choose a coin index $I\in[d]$ based on its internal randomness, previous coin choices, and outcomes. Conditioned on the choice $I$, the algorithm receives an outcome $X\sim\Bern(p_I)$ that is independent of the algorithm's history. For $i\in[d]$, we use $N_i$ to denote the sample complexity of the $i$th coin, i.e., $N_i$ is the number of times coin $i$ was tossed. We let $N=\sum_{i=1}^{d}N_i$ denote the total number of samples the algorithm has used across all $d$ coins. We use $Y$ to denote the output of the algorithm. To refer to this model, we say $\calA$ has sample access to coins with unknown biases $\boldp\in[0,1]^d$.

Our main focus will be on the $d$-$\CoinBiasEstimationProblem$ problem defined as follows.
\begin{definition}[$(d,\varepsilon,\delta)$-$\CoinBiasEstimationProblem$]
Let $d\in\mathbb{N}$ and $\varepsilon,\delta>0$. Given sample access to $d$ coins with unknown biases $\boldp=(p_1,\ldots, p_d)\in[0,1]^d$, the goal is to output an estimate $\hat\boldp\in\mathbb{R}^d$ such that
\[\Pr(\left\lVert\hat\boldp-\boldp\right\rVert_\infty\leq\varepsilon)\geq1-\delta\]
The randomness is over coin tosses and the algorithm's internal randomness.
\end{definition}

An upper bound for the $(d,\varepsilon,\delta)$-$\CoinBiasEstimationProblem$ follows directly from Chernoff Bound: sampling each coin $\widetilde{\Theta}\left(\frac{1}{\varepsilon^2}\right)$ times and outputting the empirical means gives a total sample complexity of $\widetilde{O}\left(\frac{d}{\varepsilon^2}\right)$. We state a corresponding lower bound. Interested readers may see \cite{KCG2016, Sca2024}.
\begin{lemma}\label{lem:coinbiasestimationproblemsamplecomplexity}
Let $\varepsilon,\delta\in(0,\frac{1}{8})$. Every algorithm for the $(d,\varepsilon,\delta)$-$\CoinBiasEstimationProblem$ requires $N=\Omega\left(\frac{d}{\varepsilon^2}\right)$ samples.
\end{lemma}

We often drop $(\varepsilon,\delta)$ from the notation when they are clear from context and simply refer to it as $d$-$\CoinBiasEstimationProblem$. 

\subsection{List Replicable Algorithms}
We now define the notion of list replicability in the context of $d$-$\CoinBiasEstimationProblem$~\cite{DPVV2023}. 

\begin{definition}
Let $k,d\in\mathbb{N}$ and $\varepsilon,\delta>0$. An algorithm $\calA$ is $(k,\varepsilon,\delta)$-list replicable for the $d$-$\CoinBiasEstimationProblem$ if for each bias vector $\boldp=(p_1,\ldots,p_d)\in[0,1]^d$ there exists a set of outputs $L_\boldp$ such that 
\begin{enumerate}
\item $|L_\boldp|\leq k$
\item $\forall\boldy\in L_\boldp\ldotp \lVert\boldy-\boldp\rVert_{\infty}\leq\varepsilon$
\item $\Pr(Y\in L_\boldp)\geq1-\delta$
\end{enumerate}
where $Y$ is the random variable denoting the output of $\calA$ when it has sample access to coins with bias vector $\boldp$. For a fixed algorithm, $L_\boldp$ depends only on $\boldp$ and not the samples or the algorithm's internal randomness.
\end{definition}




A $2^d$-list replicable algorithm is straightforward: estimate each bias to within $\frac{\varepsilon}{2}$, then round to the nearest point on an $\varepsilon$-spaced grid. This uses $\widetilde{O}\left(\frac{d}{\varepsilon^2}\right)$ total samples. The list size, however, can be reduced dramatically. We state the relevant results below.

\begin{theorem}[Theorem 5.2 of \cite{DPVV2023}]
Let $d\in\mathbb{N}$. For every $\varepsilon,\delta\in(0,1)$, there exists a $(d+1,\varepsilon,\delta)$-list replicable algorithm for the $d$-$\CoinBiasEstimationProblem$ using $N=O\left(\frac{d^3}{\varepsilon^2}\log\frac{d}{\delta}\right)$ samples.
\end{theorem}
\begin{theorem}[Corollary 3.3 of \cite{WDPRV2024}]
Let $d,k\in\mathbb{N}$ such that $k\in[d+1,2^d]$ and $\varepsilon,\delta\in(0,1)$. There exists a $(k,\varepsilon,\delta)$-list replicable algorithm for the $d$-$\CoinBiasEstimationProblem$ that uses $N=O\left(\frac{d^3\log^2 d}{\varepsilon^2\log^2 k}\log\frac{d}{\delta}\right)$ samples.
\end{theorem}
As mentioned earlier, we would like the list size to be $k=1$ to ensure ``perfect replicability'', i.e., for each bias vector $\boldp\in[0,1]^d$ the algorithm outputs some fixed $\varepsilon$-close output $\boldy_\boldp\in\mathbb{R}^d$ with probability at least $1-\delta$. However, such a strong notion of replicability is impossible; any $k<d+1$ is impossible, as shown in \cite{DPVV2023}.
\begin{theorem}[Theorem 5.5 of \cite{DPVV2023}]
Let $d,k\in\mathbb{N}$. For $\varepsilon\in\left(0,\frac{1}{2}\right)$, $\delta\in\left(0,\frac{1}{d+2}\right)$ and $k<d+1$, there is no $(k,\varepsilon,\delta)$-list replicable algorithm for the $d$-$\CoinBiasEstimationProblem$.
\end{theorem}

These results motivate our focus on the sample complexity of $(d+1,\varepsilon,\delta)$-list replicable algorithms and their generalizations to $(k,\varepsilon,\delta)$-list replicable algorithms.

We focus on list sizes $k\in[d+1,2^d]$, since lists of size $k<d+1$ are impossible, while for $k\geq2^d$, the sample complexity asymptotically matches that of $(d,\varepsilon,\delta)$-$\CoinBiasEstimationProblem$.

\section{Contributions}
We establish tight upper and lower bounds on the sample complexity of $(d+1,\varepsilon,\delta)$-list replicable algorithms for the $d$-$\CoinBiasEstimationProblem$. We also generalize the result and show matching upper and lower bounds for $(k,\varepsilon,\delta)$-list replicable algorithms for the $d$-$\CoinBiasEstimationProblem$ for any $k\in[d+1,2^d]$.

Our lower bound results can be stated as follows; the proof of which can be found in \Cref{sec:lowerbound}.

\begin{theorem}[\Cref{thm:lowerboundmaintheorem} Restated]
Let $\varepsilon,\delta\in(0,1/8)$. For sufficiently large $d\in\mathbb{N}$, every $(d+1,\varepsilon,\delta)$-list replicable algorithm for the $d$-$\CoinBiasEstimationProblem$ requires $N=\Omega\left(\frac{d^2}{\varepsilon^2\log d}\right)$ samples.
\end{theorem}

\begin{theorem}[\Cref{thm:lowerboundmaintheoremklist} Restated]
Let $\varepsilon,\delta\in(0,1/8)$ and $d,k\in\mathbb{N}$. If $k\in[d+1,2^d]$ then every $(k,\varepsilon,\delta)$-list replicable algorithm for the $d$-$\CoinBiasEstimationProblem$ requires $N=\Omega\left(\frac{d^2}{\varepsilon^2\log k}\right)$ samples.
\end{theorem}

For simplicity of analysis, we first prove the lower bound for algorithms when the number of tosses of each coin is fixed in advance. In \Cref{sec:generallowerbound}, we remove that assumption and prove the same lower bound for arbitrary randomized sampling-based algorithms.

Our upper bound results can be stated as follows; the proof of which can be found in \Cref{sec:upperbound}. In the statements below, we do not attempt to optimize polylogarithmic factors.


\begin{theorem}[\Cref{thm:upperboundmaintheorem} Restated]
Let $d\in\mathbb{N}$ and $\varepsilon,\delta\in(0,1)$. There exists a $(d+1,\varepsilon,\delta)$-list replicable algorithm for the $d$-$\CoinBiasEstimationProblem$ that uses $N=O\left(\frac{d^2}{\varepsilon^2}\log\frac{d}{\delta}\log\frac{1}{\delta}\right)$ samples.
\end{theorem}

\begin{theorem}[\Cref{thm:upperboundmaintheoremklist} Restated]
Let $d,k\in\mathbb{N}$ such that $k\in[d+1,2^d]$ and $\varepsilon,\delta\in(0,1)$. There exists a $(k,\varepsilon,\delta)$-list replicable algorithm for the $d$-$\CoinBiasEstimationProblem$ that uses $N=O\left(\frac{d^2\log d}{\varepsilon^2\log k}\log^2\frac{d}{\delta}\right)$ samples.
\end{theorem}

The following two tables summarize our contributions.
\begin{table}[H]
\centering
\renewcommand{\arraystretch}{1.4}
\begin{tabular}{lcc}
\toprule
Reference & Lower bound on $N$ & Upper bound on $N$ \\
\midrule
\cite{DPVV2023}
& ---
& $\displaystyle O\left(\frac{d^3}{\varepsilon^2}\log\frac{d}{\delta}\right)$ \\
This work
& $\displaystyle\Omega\left(\frac{d^2}{\varepsilon^2\log d}\right)$
& $\displaystyle O\left(\frac{d^2}{\varepsilon^2}\log\frac{d}{\delta}\log\frac{1}{\delta}\right)$ \\
\bottomrule
\end{tabular}
\caption{Sample complexity $N$ of $(d+1,\varepsilon,\delta)$-list replicable $d$-$\CoinBiasEstimationProblem$. Our work is tight up to polylogarithmic factors. \Cref{thm:upperboundmaintheorem,thm:lowerboundmaintheorem}.}
\label{tab:contributions}
\end{table}

\begin{table}[H]
\centering
\renewcommand{\arraystretch}{1.4}
\begin{tabular}{lcc}
\toprule
Reference & Lower bound on $N$ & Upper bound on $N$ \\
\midrule
\cite{WDPRV2024}
& ---
& $\displaystyle O\left(\frac{d^3\log^2 d}{\varepsilon^2\log^2 k}\log\frac{d}{\delta}\right)$ \\
This work
& $\displaystyle\Omega\left(\frac{d^2}{\varepsilon^2\log k}\right)$
& $\displaystyle O\left(\frac{d^2\log d}{\varepsilon^2\log k}\log^2\frac{d}{\delta}\right)$ \\
\bottomrule
\end{tabular}
\caption{Sample complexity $N$ of $(k,\varepsilon,\delta)$-list replicable $d$-$\CoinBiasEstimationProblem$ where $k\in[d+1,2^d]$. Our work is tight up to polylogarithmic factors. \Cref{thm:upperboundmaintheoremklist,thm:lowerboundmaintheoremklist}.}
\label{tab:contributionsklist}
\end{table}

\section{Lower Bound}\label{sec:lowerbound}
In \Cref{subsec:theneighborhoodkkmspernerlemma,subsec:lowerboundproofoverview,subsec:changeofmeasureforcointosses}, we cover preliminary results used to prove the lower bound and an overview of the proof of the lower bound. In \Cref{subsec:lowerboundford1listreplicablealgorithms}, we prove that every $(d+1,\varepsilon,\delta)$-list replicable algorithms requires $N=\Omega\left(\frac{d^2}{\varepsilon^2\log d}\right)$. Then in \Cref{subsec:lowerboundforklistreplicablealgorithms}, we generalize it to every $(k,\varepsilon,\delta)$-list replicable algorithms requires $N=\Omega\left(\frac{d^2}{\varepsilon^2\log k}\right)$ samples.
The main tool we use is the neighborhood Sperner/KKM lemma \cite{WDPRV2024}, which we describe next.

\subsection{The Neighborhood KKM/Sperner Lemma}\label{subsec:theneighborhoodkkmspernerlemma}
\begin{definition}[Sperner/KKM Coloring]
For a set of colors $C$, a coloring is a mapping $\chi\colon[0,1]^d\to C$. For a color $c\in C$, let $X_c$ denote the set of points assigned color $c$ by $\chi$, i.e., $X_c=\chi^{-1}(c)$. A coloring $\chi$ is a Sperner/KKM coloring if no two points from opposite faces of the cube get the same color, i.e., for every $c\in C$ the set $X_c$ has the property that for each coordinate $i\in[d]$, the projection $\pi_i(X_c)=\{p_i:(p_1,\ldots,p_d)\in X_c\}$ does not contain both $0$ and $1$.

An analogous definition applies when we are coloring $[a,b]^d$ where $a<b$. Still requiring that no two points from opposite faces of the cube get the same color, i.e., for every $c\in C$ the set $X_c$ has the property that for each coordinate $i\in[d]$, the projection $\pi_i(X_c)=\{p_i:(p_1,\ldots,p_d)\in X_c\}$ does not contain both $a$ and $b$.
\end{definition}

In \cite{WDPRV2024}, the authors proved the Neighborhood Sperner/KKM Lemma for colorings of $[0,1]^d$. We state the corresponding version for $[a,b]^d$ obtained by translating and rescaling the cube.
\begin{theorem}[Neighborhood Sperner/KKM Lemma \cite{WDPRV2024}]\label{thm:neighborhoodspernerkkm}
Let $a<b$ and $s=b-a$. Given a Sperner/KKM coloring of $[a,b]^d$ for any $r\in(0,\frac{s}{2}]$, there exists a point $\boldp\in[a,b]^d$ such that $\calB_{\infty}^{\circ}(r,\boldp)$ contains at least $\left(1+\frac{2}{3}\cdot\frac{r}{s}\right)^d$ points with distinct colors.
\end{theorem}

\subsection{Proof Overview}\label{subsec:lowerboundproofoverview}
We now provide a high-level overview of the lower-bound proof. Let $\calA$ be a $(d+1,\varepsilon,\delta)$-list replicable algorithm for $d$-$\CoinBiasEstimationProblem$ and $Y$ be the random variable denoting the output of $\calA$. For simplicity, we assume that $\calA$ does not use any internal randomness and the sample complexity is $N$.

Given $\calA$, we color the region $\calR=\left[\frac{1}{2}-2\varepsilon,\frac{1}{2}+2\varepsilon\right]^d$ as follows: for a bias vector $\boldp\in\calR$, we color it with the most likely output of $\calA$ when given sample access to $\boldp$, i.e., color $\boldp$ with $\argmax_{\boldy\in L_\boldp}\Pr\nolimits_P(Y=\boldy)$, where $P$ is the distribution that algorithm $\calA$ observes when given sample access to $\boldp$. We first observe that this is a valid Sperner/KKM coloring. By the neighborhood Sperner/KKM lemma, for every radius $r\in(0,2\varepsilon]$ there exists a point $\boldp\in\calR$ such that there are at least $M:=\left(1+\frac{r}{6\varepsilon}\right)^d$ distinct colors in $\calB_\infty^\circ(r,\boldp)$. This means that the algorithm must change its most likely output many times in this small region. We prove that for this to happen, the sample complexity of the algorithm must be high. 

Let $\boldp\in\calR$ be the point from the Neighborhood Sperner/KKM Lemma, and let $\boldq_1, \cdots, \boldq_M\in\calB_\infty^\circ(r,\boldp)$ be points with distinct colors $\boldy_1,\ldots,\boldy_M$. This means that the most likely output of $\calA$ when given sample access to $\boldp_i$ is $\boldy_i$. Let $E_i$ be the event that the output $Y$ is $\boldy_i$. Let $P$ be the distribution that algorithm $\calA$ observes when given sample access to $\boldp$, similarly $Q_i$ is the distribution $\calA$ observes when given sample access to $\boldq_i$

Now, we want to look at the probability of $E_i$ when $\calA$ has sample access to $\boldp$, let's denote this with $\Pr_{P}(E_i)$. We can relate this to the probability of $E_i$ when $\calA$ has sample access to $\boldq_i$, we denote this with $\Pr_{Q_i}(E_i)$. Our first approach was to relate them in an additive manner. Note that 
$\Pr\nolimits_P(E_i)\geq \Pr_{Q_i}(E_i)-\dTV(P,Q_i
)$, and by Pinsker's inequality $\Pr\nolimits_P(E_i)\geq \Pr_{Q_i}(E_i)-\sqrt{\frac{1}{2}\DKL(P\parallel Q_i)}$. 
Since the sample complexity of algorithm is $N$, this implies that $\Pr\nolimits_P(E_i)\geq \Pr_{Q_i}(E_i)-\sqrt{\frac{8}{3}N\lVert \boldp-\boldq_i\rVert_\infty^2}$. So, we obtain the following inequality
\[1 \geq\sum_{i=1}^M \Pr\nolimits_P(E_i)\geq \sum_{i=1}^M \Pr\nolimits_{Q_i}(E_i)- \sum_{i=1}^M \sqrt{\frac{8}{3}N\lVert \boldp - \boldq_i \rVert_\infty^2}\]

Note that $\Pr\nolimits_{Q_i}(E_i) \geq \frac{1-\delta}{d+1}$ (since most likely output has probability at least $\frac{1-\delta}{d+1}$), and $\lVert \boldp - \boldq_i \rVert \leq r$,  From this, we obtain (ignoring the constant $8/3$)
\[1 \geq \frac{M(1-\delta)}{d+1} - Mr\sqrt{N}\]
When we work out the details and solve the above inequality for $N$, we do not obtain any non-trivial lower bound on $N$. The main issue is that the inequality $\Pr\nolimits_P(E_i)\geq \Pr_{Q_i}(E_i)-\sqrt{\frac{8}{3}N\lVert \boldp-\boldq_i\rVert_\infty^2}$ is too weak for our setting. Instead, we use a multiplicative version of the inequality $\Pr\nolimits_{P}(E_i)\geq \left(\Pr\nolimits_{Q_i}(E)\right)^2e^{-\frac{16}{3}N\lVert\boldp-\boldq_i\rVert_\infty^2}$, whose details we describe next.


\subsection{Change of Measure for Coin Tosses}\label{subsec:changeofmeasureforcointosses}
This section covers some rudimentary properties of probability theory. We prove them for the sake of completeness. Interested readers may look at \cite{LSS2014,Mir2017,Mah2026}.
\begin{lemma}\label{lem:expectedsquaredderivativeproductlemma}
Let $P=R_1\otimes \ldots\otimes R_n$ and $Q=S_1\otimes \ldots\otimes S_n$ be product distributions. If $\forall i\in[n]\ldotp S_i\ll R_i$ then
\[\EE_P\left[\left(\frac{dQ}{dP}\right)^2\right]=\prod_{i=1}^{n}\EE_{R_i}\left[\left(\frac{dS_i}{dR_i}\right)^2\right]\]
\end{lemma}
\begin{proof}
We prove the case $n=2$, and for simplicity we assume a discrete sample space. By using the definition of expectation and the fact that $P=R_1\otimes R_2$ and $Q=S_1\otimes S_2$ are product distributions, we have
\begin{align*}
\EE_P\left[\left(\frac{dQ}{dP}\right)^2\right]&=\sum_{x_1,x_2}\left(\frac{Q(x_1,x_2)}{P(x_1,x_2)}\right)^2 P(x_1,x_2)\\
&=\sum_{x_1}\sum_{x_2}\left(\frac{S_1(x_1)S_2(x_2)}{R_1(x_1)R_2(x_2)}\right)^2 R_1(x_1)R_2(x_2)\\
&=\left[\sum_{x_1}\left(\frac{S_1(x_1)}{R_1(x_1)}\right)^2 R_1(x_1)\right]\left[\sum_{x_2}\left(\frac{S_2(x_2)}{R_2(x_2)}\right)^2 R_2(x_2)\right]\\
&=\EE_{R_1}\left[\left(\frac{dS_1}{dR_1}\right)^2\right]\EE_{R_2}\left[\left(\frac{dS_2}{dR_2}\right)^2\right]
\end{align*}
The general case follows by induction on $n$.
\end{proof}

\begin{lemma}\label{lem:expectedsquaredderivatebernoullibound}
Let $p,q\in(0,1)$. If $R=\Bern(p)$ and $S=\Bern(q)$ then
\[\EE_R\left[\left(\frac{dS}{dR}\right)^2\right]=1+\frac{(p-q)^2}{p(1-p)}\]
In particular, if $\frac{1}{4 }\leq p\leq\frac{3}{4}$ then
\[\EE_R\left[\left(\frac{d S}{dR}\right)^2\right]\leq e^{\frac{16}{3}(p-q)^2}\]
\end{lemma}
\begin{proof}
\begin{align*}
\EE_R\left[\left(\frac{dS}{dR}\right)^2\right]&=p\cdot\left(\frac{q}{p}\right)^2+(1-p)\cdot\left(\frac{1-q}{1-p}\right)^2\\
&=\frac{q^2}{p}+\frac{(1-q)^2}{1-p}\\
&=\frac{q^2(1-p)+p(1-q)^2}{p(1-p)}\\
&=\frac{q^2+p-2qp}{p(1-p)}\\
&=\frac{p-p^2}{p(1-p)}+\frac{(p-q)^2}{p(1-p)}\\
&=1+\frac{(p-q)^2}{p(1-p)}
\end{align*}
In particular, if $p\in\left[\frac{1}{4},\frac{3}{4}\right]$ then due to convexity of $\frac{1}{p(1-p)}$ we have that $\frac{1}{p(1-p)}\leq\frac{16}{3}$ and therefore
\begin{align*}
\EE_R\left[\left(\frac{dS}{dR}\right)^2\right]&\leq1+\frac{16}{3}(p-q)^2\\
&\leq e^{\frac{16}{3}(p-q)^2}\tag{$1+x\leq e^x$}
\end{align*}
This concludes the proof.
\end{proof}

Let $\boldp,\boldq\in[0,1]^d$, and also let $N_1,\ldots,N_d\in\mathbb{N}$ with $N=\sum_{i=1}^{d}N_i$. Here $N_i$ denotes the number of times coin $i$ was tossed, and $N$ denotes the total number of tosses across all coins. Define the distributions
\begin{align*}
P&=\bigotimes_{i=1}^{d}(\Bern(p_i)^{\otimes N_i})\\
Q&=\bigotimes_{i=1}^{d}(\Bern(q_i)^{\otimes N_i})
\end{align*}
Thus $P,Q$ are distributions on $\{0,1\}^N$ where the first $N_1$ bits are from the first coin, the next $N_2$ bits are from the second coin, and so on.
\begin{lemma}\label{lem:squaredderivativeboundfordproductbernoulli}
If $\mathbf{\frac{1}{4}}\leq\boldp\leq\mathbf{\frac{3}{4}}$ then $\EE_P\left[\left(\frac{dQ}{dP}\right)^2\right]\leq e^{\frac{16}{3}N\left\lVert\boldp-\boldq\right\rVert_{\infty}^2}$
\end{lemma}
\begin{proof}
By using \Cref{lem:expectedsquaredderivativeproductlemma,lem:expectedsquaredderivatebernoullibound} we have
\begin{align*}
\EE_P\left[\left(\frac{dQ}{dP}\right)^2\right]&=\prod_{i=1}^{d}\left(1+\frac{(p_i-q_i)^2}{p_i(1-p_i)}\right)^{N_i}\\
&\leq\prod_{i=1}^{d}e^{\frac{16}{3}N_i(p_i-q_i)^2}\\
&=e^{\frac{16}{3}\sum_{i=1}^{d}N_i(p_i-q_i)^2}\\
&\leq e^{\frac{16}{3}\sum_{i=1}^{d}N_i\left\lVert\boldp-\boldq\right\rVert_{\infty}^2}\tag{$\forall i\ldotp |p_i-q_i|\leq\left\lVert\boldp-\boldq\right\rVert_{\infty}$}\\
&=e^{\frac{16}{3}N\left\lVert\boldp-\boldq\right\rVert_{\infty}^2}\tag*{\qedhere}
\end{align*}
\end{proof}

\begin{lemma}\label{lem:eventclosenessbound}
If $\mathbf{\frac{1}{4}}\leq\boldp\leq\mathbf{\frac{3}{4}}$ then for any measurable event $E$
\[\Pr\nolimits_{P}(E)\geq \left(\Pr\nolimits_{Q}(E)\right)^2e^{-\frac{16}{3}N\lVert\boldp-\boldq\rVert_\infty^2}\]
In particular, if $\lVert\boldp-\boldq\rVert_\infty\leq r$ then
\[\Pr\nolimits_{P}(E)\geq \left(\Pr\nolimits_{Q}(E)\right)^2e^{-\frac{16}{3}Nr^2}\]
\end{lemma}
\begin{proof}
By using properties of expectation and probability, we have
\begin{align*}
\left(\Pr\nolimits_{Q}(E)\right)^2 &=\left(\EE_Q[\mathbf{1}_{\{E\}}]\right)^2\\
                &=\left(\EE_P\left[\mathbf{1}_{\{E\}}\frac{dQ}{dP}\right]\right)^2\\
                &\leq\EE_P\left[\mathbf{1}_{\{E\}}^2\right]\EE_P\left[\left(\frac{dQ}{dP}\right)^2\right]\tag{Cauchy-Schwarz}\\
                &=\EE_P\left[\mathbf{1}_{\{E\}}\right]\EE_P\left[\left(\frac{dQ}{dP}\right)^2\right]\\
                &=\Pr\nolimits_{P}\left(E\right)\EE_P\left[\left(\frac{dQ}{dP}\right)^2\right]\\
                &\leq\Pr\nolimits_{P}\left(E\right)e^{\frac{16}{3}N\left\lVert\boldp-\boldq\right\rVert_{\infty}^2}\tag{\Cref{lem:squaredderivativeboundfordproductbernoulli}}
\end{align*}
Rearranging the above gives us the desired bound, and substituting $\lVert\boldp-\boldq\rVert_\infty$ with $r$ gives us the second bound.
\end{proof}

\subsection{\texorpdfstring{Lower Bound for $(d+1)$-List Replicable Algorithms}{Lower Bound for (d+1)-List Replicable Algorithms}}\label{subsec:lowerboundford1listreplicablealgorithms}
\begin{theorem}\label{thm:lowerboundmaintheorem}
Let $\varepsilon,\delta\in(0,1/8)$. For sufficiently large $d\in\mathbb{N}$, every $(d+1,\varepsilon,\delta)$-list replicable algorithm for the $d$-$\CoinBiasEstimationProblem$ requires $N=\Omega\left(\frac{d^2}{\varepsilon^2\log d}\right)$ samples.
\end{theorem}
For simplicity's sake, we assume that the algorithm is deterministic and the sample complexity of each coin $i\in[d]$ is fixed to be $N_i\in\mathbb{N}$ in advance, i.e., the algorithm receives as input a binary string of length $N=\sum_{i=1}^{d}N_i$, where the first $N_1$ bits are i.i.d. Bernoulli tosses according to the first coin, and the next $N_2$ bits are i.i.d. Bernoulli tosses according to the second coin, and so on. We prove the lower bound against arbitrary algorithms with sample access to $d$ coins in \Cref{sec:generallowerbound}.

Let $\calA\colon\{0,1\}^{N}\to\calY$ be a deterministic algorithm that is $(d+1,\varepsilon,\delta)$-list replicable algorithm such that each coin $i\in[d]$ is tossed exactly $N_i$ with total sample complexity of $N=\sum_{i=1}^{d}N_i$. Let $Y$ be the random variable denoting the output of this algorithm. Because $\calA$ is list replicable, for each bias vector $\boldp\in[0,1]^d$, there exists a set $L_\boldp\subseteq\calY$ of size $|L_\boldp|\leq d+1$ such that each $\boldy\in L_\boldp$ it holds that $\lVert\boldp-\boldy\rVert_\infty\leq\varepsilon$ and
\[\Pr\nolimits_P(Y\in L_\boldp)\geq1-\delta\]
We let $\boldy_\boldp=\argmax_{\boldy\in L_\boldp}\Pr\nolimits_P(Y=\boldy)$ denote the most likely output from $L_\boldp$, breaking ties with some fixed rule. Using an averaging argument, the most likely output must appear with probability at least $\frac{1-\delta}{|L_\boldp|}$; otherwise, the sum of probabilities of outputs in $L_\boldp$ would fail to add up to $1-\delta$. Therefore
\[\Pr\nolimits_P(Y=\boldy_\boldp)\geq\frac{1-\delta}{|L_\boldp|}\geq\frac{7}{8(d+1)}\geq\frac{1}{2(d+1)}=:\alpha\]
The second inequality was obtained by using $|L_\boldp|\leq d+1$ and $\delta<\frac{1}{8}$.

If an output $\boldy$ is $\varepsilon$-close to $\boldp$, and the distance between $\boldp$ and $\boldq$ is $>2\varepsilon$ then $\boldy$ cannot be $\varepsilon$-close to $\boldq$. In particular, this must also apply to the most likely output of $L_\boldp$ formalized below.
\begin{lemma}\label{lem:farpointscannotsharemostlikelyoutput}
Let $\boldp,\boldq\in[0,1]^d$. If $\lVert\boldp-\boldq\rVert_\infty>2\varepsilon$ then $\boldy_\boldp\notin L_\boldq$.
\end{lemma}
\begin{proof}
Recall, by the definition of list replicability, we require that $L_\boldp$ consist only of points that are $\varepsilon$-close to $\boldp$, i.e., if $\boldy\in L_\boldp$ then $\lVert\boldp-\boldy\rVert_\infty\leq\varepsilon$. In particular, this must also hold for $\boldy_\boldp$, yielding $\lVert\boldp-\boldy_\boldp\rVert_\infty\leq\varepsilon$. Using triangle inequality, we have
\[\lVert\boldp-\boldy_\boldp\rVert_\infty+\lVert\boldy_\boldp-\boldq\rVert_\infty\geq\lVert\boldp-\boldq\rVert_\infty\]
Substituting $\lVert\boldp-\boldq\rVert_\infty>2\varepsilon$ and $\lVert\boldp-\boldy_\boldp\rVert_\infty\leq\varepsilon$ and rearranging yields $\lVert\boldy_\boldp-\boldq\rVert_\infty>\varepsilon$. Thus $\boldy_\boldp$ is not $\varepsilon$-close to $\boldq$. Therefore $\boldy_\boldp\notin L_\boldq$.
\end{proof}

We consider the region $\calR=\left[\frac{1}{2}-2\varepsilon,\frac{1}{2}+2\varepsilon\right]^d\subseteq\left[\frac{1}{4},\frac{3}{4}\right]^d$ and we color the region $\calR$ with the coloring $\chi\colon\calR\to\calY$ defined as 
\[\chi(\boldp)=\boldy_\boldp\]

\begin{lemma}
$\chi$ is a Sperner/KKM coloring of $\calR$.
\end{lemma}
\begin{proof}

We now prove the Sperner/KKM coloring property. Consider any two points on opposite faces $\boldp,\boldq\in\calR$. Since they are on opposite faces, there exists a coordinate $i\in[d]$ such that $|p_i-q_i|=4\varepsilon$. In particular, this implies that $\lVert\boldp-\boldq\rVert_\infty\geq4\varepsilon>2\varepsilon$. Via \Cref{lem:farpointscannotsharemostlikelyoutput} we have that $\chi(\boldp)=\boldy_\boldp\notin L_\boldq$. In particular, $\chi(\boldp)=\boldy_\boldp\neq \boldy_\boldq=\chi(\boldq)$.
\end{proof}

Since $\chi$ is a valid Sperner/KKM coloring, the Neighborhood Sperner/KKM, \Cref{thm:neighborhoodspernerkkm}, applies with $a=\frac{1}{2}-2\varepsilon$, $b=\frac{1}{2}+2\varepsilon$ and $s=b-a=4\varepsilon$. Let $r\in\left(0,2\varepsilon\right]$ to be decided later, and $\boldp\in[a,b]^d$ be the point such that $\calB_{\infty}^{\circ}(r,\boldp)$ that contains at least $M:=\left(1+\frac{r}{6\varepsilon}\right)^d$ colors. Let $C:=\left\{\boldy_\boldq:\boldq\in\calB_{\infty}^{\circ}(r,\boldp)\cap\calR\right\}$ be the set of colors used to color $\calB_{\infty}^{\circ}(r,\boldp)\cap\calR$ and for each $c\in C$ let $\boldq_c$ be some point in $\calB_{\infty}^{\circ}(r,\boldp)\cap\calR$ for which $\chi(\boldq_c)=c$.

\begin{lemma}\label{lem:lowerboundmainlemma}
$\frac{16}{3}Nr^2\geq \frac{dr}{10\varepsilon}-2\ln(2(d+1))$
\end{lemma}
\begin{proof}
For any color $c\in C$, let $\boldq_c\in\calB_{\infty}^{\circ}(r,\boldp)$ be as described above, so we have $\lVert \boldp-\boldq_c\rVert_\infty<r$. We invoke \Cref{lem:eventclosenessbound} with the event $E=\{Y=c\}$ yielding
\[\Pr\nolimits_P(Y=c)\geq\left(\Pr\nolimits_{Q_c}(Y=c)\right)^2e^{-\frac{16}{3}Nr^2}\geq\alpha^2e^{-\frac{16}{3}Nr^2}\]
The last inequality follows from our definition of the coloring $\chi$, we colored $\boldq_c$ with the most likely output $\boldy_{\boldq_c}$. Therefore $\Pr\nolimits_{Q_c}(Y=c)=\Pr\nolimits_{Q_c}(Y=\boldy_{\boldq_c})\geq\alpha$.

Summing over all possible colors, of which there are at least $M$ many, and because $\Pr\nolimits_P$ is a probability measure, we have the following
\[M\alpha^2e^{-\frac{16}{3}Nr^2}\leq \sum_{c\in C}\Pr\nolimits_P(Y=c)\leq 1\]
Taking $\ln$ on both sides and rearranging and simplifying yields 
\[\frac{16}{3}Nr^2\geq d\ln\left(1+\frac{r}{6\varepsilon}\right)-2\ln\frac{1}{\alpha}\]
Using the fact that $x\in(0,1)\implies \ln(1+x)\geq x\ln 2$ and $r\in(0,2\varepsilon]$, thus $\frac{r}{6\varepsilon}\in(0,\frac{1}{3}]$ we have
\[\ln\left(1+\frac{r}{6\varepsilon}\right)\geq \frac{r}{6\varepsilon}\ln 2\geq\frac{r}{10\varepsilon}\]
Substituting $\alpha=\frac{1}{2(d+1)}$ yields
\[\frac{16}{3}Nr^2\geq \frac{dr}{10\varepsilon}-2\ln(2(d+1))\qedhere\]
\end{proof}

Now pick $r=\frac{40\varepsilon}{d}\ln(2(d+1))$ so that $\frac{dr}{10\varepsilon}=4\ln(2(d+1))$, for sufficiently large $d$, we have $r\leq 2\varepsilon$ as $40\ln(2(d+1))\ll d$. Substituting $r$ and simplifying the expression yields
\[N\geq\frac{3}{12800}\cdot\frac{d^2}{\varepsilon^2\ln(2(d+1))}=\Omega\left(\frac{d^2}{\varepsilon^2\log d}\right)\]

This concludes the proof for \Cref{thm:lowerboundmaintheorem} for the case of deterministic algorithms whose sample complexities are fixed in advance. Extending it to arbitrary randomized sampling-based algorithms is mostly technical; we need to inspect the distribution on the transcript $(W,I_1,X_1,\ldots,I_N,X_N)$, where $W$ is the internal randomness of the algorithm, for $t\in[N]$, $I_t$ is the coin selected at time $t$ and $X_t$ is the outcome of the toss at time $t$. Bias vectors $\boldp,\boldq\in[0,1]^d$ induce probability distributions $P,Q$ on the transcript as described above. If $\boldp\in\left[\frac{1}{4},\frac{3}{4}\right]^d$ then the relation in \Cref{lem:squaredderivativeboundfordproductbernoulli} also holds in this case $\EE_P\left[\left(\frac{dQ}{dP}\right)^2\right]\leq e^{\frac{16}{3}N\left\lVert\boldp-\boldq\right\rVert_{\infty}^2}$ and remaining argument is the same.

\subsection{\texorpdfstring{Lower Bound for $k$-List Replicable Algorithms}{Lower Bound for k-List Replicable Algorithms}}\label{subsec:lowerboundforklistreplicablealgorithms}
We now generalize \Cref{thm:lowerboundmaintheorem} to a list of size $k$ as follows
\begin{theorem}\label{thm:lowerboundmaintheoremklist}
Let $\varepsilon,\delta\in(0,1/8)$ and $d,k\in\mathbb{N}$. If $k\in[d+1,2^d]$ then every $(k,\varepsilon,\delta)$-list replicable algorithm for the $d$-$\CoinBiasEstimationProblem$ requires $N=\Omega\left(\frac{d^2}{\varepsilon^2\log k}\right)$ samples.
\end{theorem}

\begin{proof}
First assume $k\leq\frac{1}{2}e^{\frac{d}{20}}$ then the argument remains similar from \Cref{sec:lowerbound}; the main bound that changes is
\[\Pr\nolimits_P(Y=\boldy_\boldp)\geq\frac{1-\delta}{|L_\boldp|}\geq\frac{7}{8k}\geq\frac{1}{2k}=:\beta\]
\Cref{lem:lowerboundmainlemma} now becomes
\[\frac{16}{3}Nr^2\geq \frac{dr}{10\varepsilon}-2\ln(2k)\]
If $k\leq\frac{1}{2}e^{\frac{d}{20}}$ then our choice $r:=\frac{40\varepsilon}{d}\ln(2k)\leq2\varepsilon$ holds and therefore
\[N\geq\frac{3}{12800}\cdot\frac{d^2}{\varepsilon^2\ln(2k)}=\Omega\left(\frac{d^2}{\varepsilon^2\log k}\right)\]

For $k>\frac{1}{2}e^{\frac{d}{20}}$, every $(k,\varepsilon,\delta)$-list replicable algorithm also solves the $(d,\varepsilon,\delta)$-$\CoinBiasEstimationProblem$. Using \Cref{lem:coinbiasestimationproblemsamplecomplexity} gives 
\[N=\Omega\left(\frac{d}{\varepsilon^2}\right)=\Omega\left(\frac{d^2}{\varepsilon^2\log k}\right)\]
The last bound was by using the fact that $k>\frac{1}{2}e^{\frac{d}{20}}$.
\end{proof}

Note that, even obtaining a list size of $k=\poly(d)$ requires $N=\Omega\left(\frac{d^2}{\varepsilon^2\log d}\right)$ samples.

\section{Upper Bound}\label{sec:upperbound}

In this section, we prove \Cref{thm:upperboundmaintheorem,thm:upperboundmaintheoremklist}. 
We first introduce the main concepts intuitively in \Cref{subsec:upperboundproofoverview} and then formalize the notions in \Cref{subsec:shiftsroundingandcanonicallist}.

\subsection{Proof Overview}\label{subsec:upperboundproofoverview}

\begin{figure}[H]
    \centering
    \resizebox{3.3in}{!}{\begin{tikzpicture}[x=2.6cm,y=2.6cm,font=\small]

\def\xstart{3.965}
\def\ystart{4.765}

\def\redwidth{0.07}
\def\orangewidth{0.18}

\def\gridsize{2.5}
\draw[step=1,gray!50,very thin]
    (2.8,3.8) grid ({2.8+\gridsize},{3.8+\gridsize});

\coordinate (A) at (\xstart,\ystart);
\coordinate (B) at ({\xstart+1},{\ystart+1});
\coordinate (C) at ({\xstart+1},\ystart);

\draw[densely dashed,gray!80]
    (A) -- node[below,text=black] {$\varepsilon$} (C)
        -- node[right,text=black] {$\varepsilon$} (B);

\draw[green!50!black,line width=2pt,line cap=butt] (A) -- (B);
\foreach \bandwidth/\bandcolor in {\orangewidth/orange,\redwidth/red} {
    \foreach \s in {\xstart,\ystart} {
        \pgfmathtruncatemacro{\nmin}{floor(\s)-1}
        \pgfmathtruncatemacro{\nmax}{ceil(\s+1)+1}
        \foreach \n in {\nmin,...,\nmax} {
            \pgfmathsetmacro{\tstart}{max(0,\n-\bandwidth-\s)}
            \pgfmathsetmacro{\tend}{min(1,\n+\bandwidth-\s)}
            \ifdim\tstart pt<\tend pt
                \draw[draw=\bandcolor,line width=2pt,line cap=butt]
                    ({\xstart+\tstart},{\ystart+\tstart}) --
                    ({\xstart+\tend},{\ystart+\tend});
            \fi
        }
    }
}

\fill (A) circle (1.3pt) node[below] {$\boldp$};
\fill (B) circle (1.3pt) node[above left] {$u=\varepsilon$};

\path (3,4) -- node[below] {$\varepsilon$} (4,4);
\path (3,4) -- node[left]  {$\varepsilon$} (3,5);
\path (4,4) -- node[below] {$\varepsilon$} (5,4);
\path (3,5) -- node[left]  {$\varepsilon$} (3,6);

\def\upos{0.50}    
\def\cixscale{1}   
\def\ciyscale{1}   

\pgfmathsetmacro{\cix}{\xstart+\upos}
\pgfmathsetmacro{\ciy}{\ystart+\upos}

\pgfmathsetmacro{\alphax}{\cixscale*abs(\cix-round(\cix))}
\pgfmathsetmacro{\alphay}{\ciyscale*abs(\ciy-round(\ciy))}

\draw[blue!70!black,densely dashed]
    ({\cix-\alphax},{\ciy-\alphay}) rectangle
    ({\cix+\alphax},{\ciy+\alphay});

\draw[->,blue!70!black]
    (\cix,\ciy) -- node[above,font=\scriptsize] {$\alpha_x$}
    ({\cix-\alphax},\ciy);
\draw[->,blue!70!black]
    (\cix,\ciy) -- node[right,font=\scriptsize] {$\alpha_y$}
    (\cix,{\ciy-\alphay});

\fill (\cix,\ciy) circle (1.3pt)
    node[above=4pt] {$\boldp'$};

\node[anchor=south east,fill=white,inner sep=2pt,font=\scriptsize]
    at (4.1,4.1) {
    \setlength{\tabcolsep}{2pt}
    \begin{tabular}{@{}ll@{}}
        \tikz{\draw[red,line width=1.5pt] (0,0)--(0.22,0);}
        & High sample cost \\
        \tikz{\draw[orange,line width=1.5pt] (0,0)--(0.22,0);}
        & Moderate sample cost \\
        \tikz{\draw[green!50!black,line width=1.5pt] (0,0)--(0.22,0);}
        & Low sample cost
    \end{tabular}
};
\end{tikzpicture}}
    \caption{Intuition of the Upper Bound Algorithm}
    \label{fig:upperboundintuition}
    \Description{A diagram illustrating the intuition of the upper bound}
\end{figure}
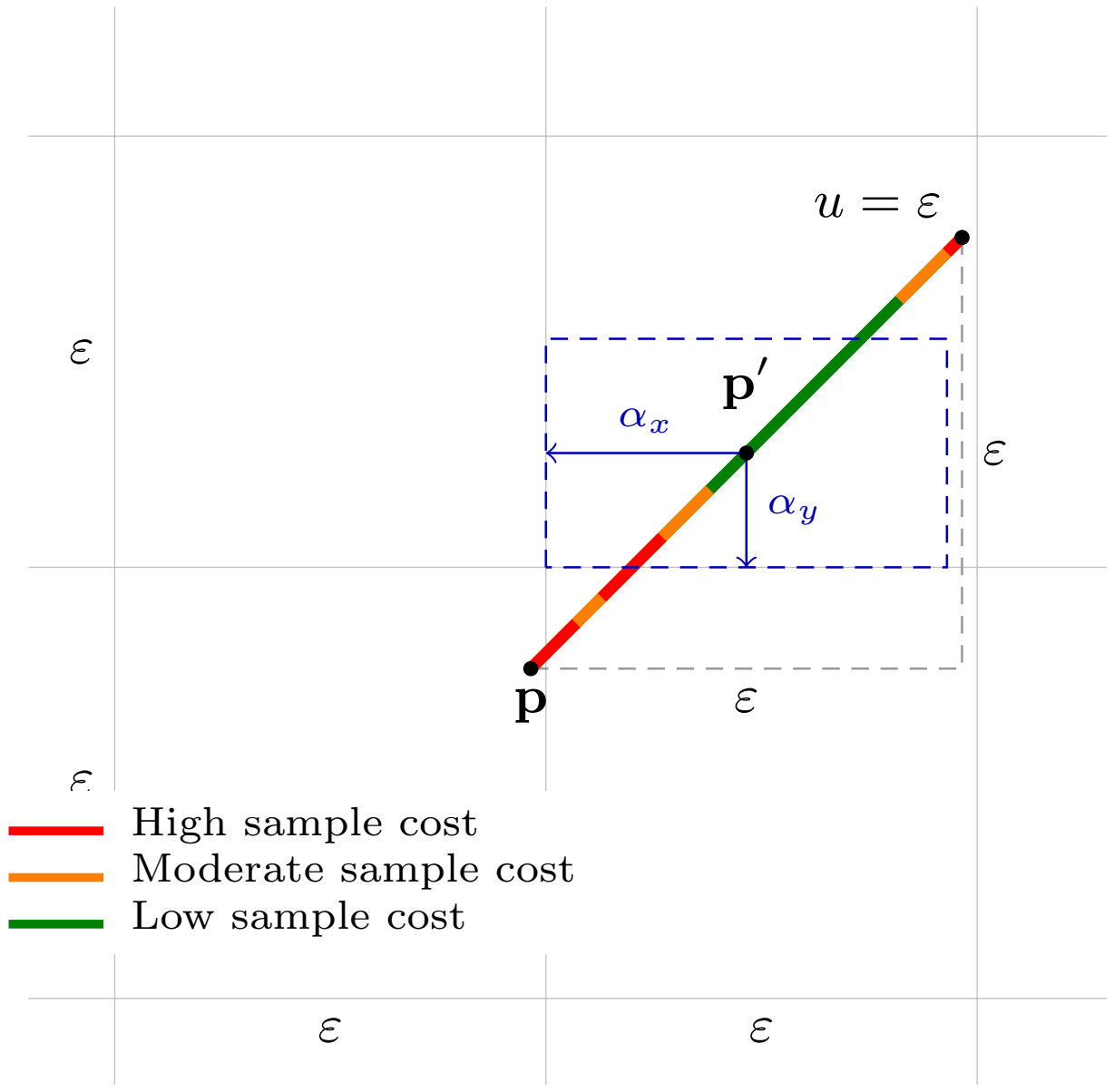

Consider a simplified setting with $d=2$ coins and an $\varepsilon$-grid as depicted in \Cref{fig:upperboundintuition}. Let $\boldp \in [0, 1]^d$ be a bias vector. A simple idea is to estimate the bias of each coin within $\frac{\varepsilon}{2}$. Then, we round the estimate to the bottom left of the $\varepsilon$-square in which it falls, this has the effect of rounding both coordinates down to the nearest integer multiple of $\varepsilon$. However, this algorithm is not 3-list replicable, because if $\boldp$ lies on a grid point, then the approximation can lie in any of the 4 adjacent $\varepsilon$-squares. To avoid this, we use the standard technique of random shifting and rounding. The technical challenge is to bound the sample complexity. 

Given $\boldq = (x, y) \in [0, 1]^2$, the $(\alpha_x, \alpha_y)$-confidence interval around $\boldq$ is the rectangle with corners from $(x \pm \alpha_x, y \pm \alpha_y)$.
Let $u\in[0,\varepsilon)$ be a shift parameter that induces a line segment of length $\varepsilon$ (in $\ell_\infty$ sense) at a $45^\circ$ angle. Note that this line intersects at most 3 cells (assuming each grid point belongs to a single cell).
We estimate the shifted vector $\boldp'=\boldp+u\cdot\mathbf{1}$ by adding $u$ to each empirical mean obtained from the original coins. We will focus on the largest confidence interval $(\alpha_x,\alpha_y)$ around $\boldp'= (x, y)$ that is strictly contained inside a {\em single} $\varepsilon$-grid cell. If we have an estimate $\hat{x}$ that is $\alpha_x$-close to $x$ and an estimate $\hat{y}$ that is $\alpha_y$-close to $y$, then all possible estimates lie in a single cell with high probability. Thus, rounding the estimate $(\hat{x}, \hat{y})$ to the bottom left of the cell yields the same value. The sample complexity required to get samples within this confidence interval with high probability is $\widetilde{O}\left(\frac{1}{\alpha_x^2}+\frac{1}{\alpha_y^2}\right)$. Since any line segment of length $\varepsilon$ at $45^\circ$ from $\boldp$ intersects at most three cells, this ensures that if we randomly pick $u$, the algorithm is $3$-list replicable (provided we know $\alpha_x$ and $\alpha_y$).

For the purposes of illustration, we classify the regions along this line into 3 regions colored based on the behavior of their confidence intervals. When $\boldp'$ lies in a green region, the required confidence interval is $\alpha_x\approx\varepsilon$ and $\alpha_y\approx\varepsilon$. In the orange region, the confidence interval needs to satisfy $\alpha_x\ll \varepsilon$ or $\alpha_y\ll \varepsilon$. Finally, in the red region, a confidence interval of $\alpha_x\approx0$ or $\alpha_y\approx0$ is required. Thus, some shifts (orange region) require higher sample complexity, whereas some shifts (green region) require fewer samples. Shifts that lead to the red region can be considered a failure of the algorithm.
Our algorithm limits the work spent on these shifts, keeping the expected sample complexity small when the shift is chosen uniformly at random.
A technicality is that our algorithm will not know $\alpha_x,\alpha_y$ a priori, but will iteratively get smaller and smaller estimates until the confidence interval lies in a single cell.

In the case of $d$ coins, our analysis shows that a uniformly at random selected $U\sim\Unif([0,\varepsilon))$ yields an expected sample complexity of $\tilde{O}(\frac{d}{\varepsilon^2})$ per coin; thus the expected total sample complexity is $\tilde{O}(\frac{d^2}{\varepsilon^2})$. List Replicability follows from the fact that this line segment intersects at most $d+1$ grid cells in $d$ dimensions. 

\subsection{Shifts, Rounding and Canonical List}\label{subsec:shiftsroundingandcanonicallist}
To round down $x\in\mathbb{R}$ to the nearest $\varepsilon$, it can be done so by the mapping $x\mapsto\lfloor\frac{x}{\varepsilon}\rfloor\varepsilon$. For example, if $x=2.5\varepsilon$ then it maps to $2\varepsilon$. So $\frac{x}{\varepsilon}$ extracts what multiple of $\varepsilon$ is $x$ (in this case $2.5$), and then we floor only the multiple part of $\varepsilon$ to the nearest integer. This flooring process reduces the output by at most $\varepsilon$.

For our algorithm, before rounding, we perturb the initial $x$ by a shift $u\in[0,\varepsilon)$ and then round down to the nearest multiple of $\varepsilon$.

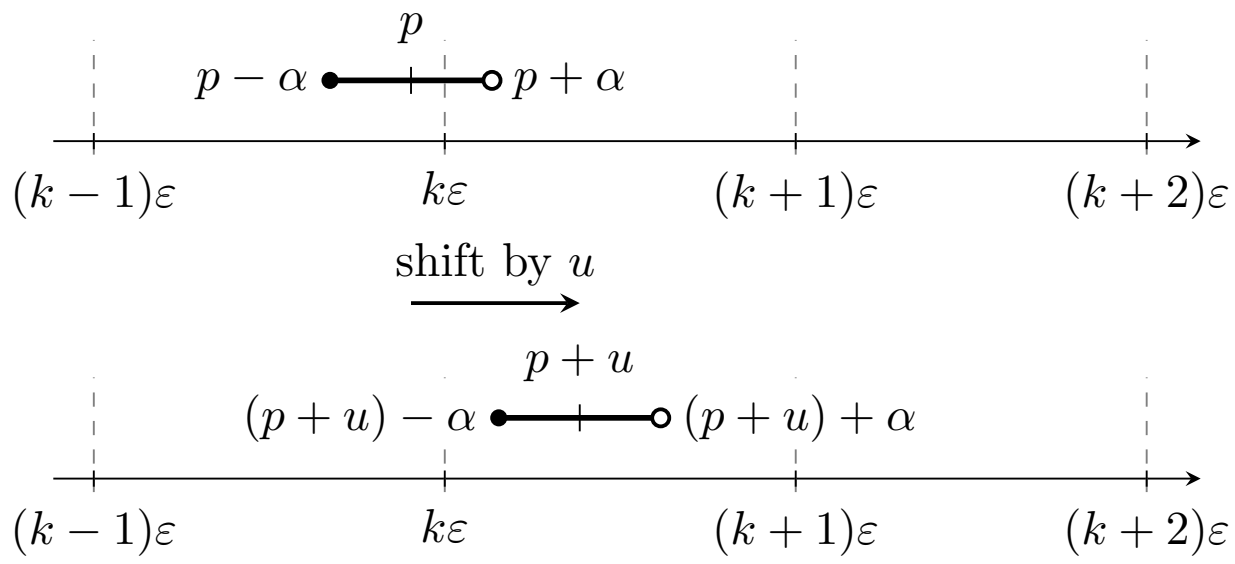
\begin{figure}[H]
    \centering
    \resizebox{3.3in}{!}{\begin{tikzpicture}[
    x=1cm,y=1cm,>=stealth,
    every node/.style={font=\normalsize},
    endpoint label/.style={fill=white,inner sep=1.5pt}
]

\def\binwidth{2.6}        
\def\intervalradius{0.6}  
\def\centerposition{2.35} 
\def\shiftamount{1.25}   

\def\rowgap{2.5}          
\def\intervalheight{0.45}
\def\titleheight{1}

\foreach \row in {0,1} {
    \begin{scope}[shift={(0,{-\row*\rowgap})}]
        \coordinate (axis-left) at (0,0);
        \coordinate (axis-right) at ({3*\binwidth},0);
        \draw[->] ([xshift=-3mm]axis-left)
            -- ([xshift=4mm]axis-right);

        \foreach \j/\label in {
            0/{(k-1)\varepsilon},1/{k\varepsilon},
            2/{(k+1)\varepsilon},3/{(k+2)\varepsilon}} {
            \coordinate (boundary) at ({\j*\binwidth},0);
            \coordinate (guide-top)
                at ({\j*\binwidth},\intervalheight);
            \draw[gray,dashed] ([yshift=-1mm]boundary)
                -- ([yshift=3mm]guide-top);
            \draw ([yshift=-0.6mm]boundary)
                -- ([yshift=0.6mm]boundary);
            \node[below=1mm] at (boundary) {$\label$};
        }
    \end{scope}
}

\coordinate (P) at (\centerposition,\intervalheight);
\coordinate (L) at ($(P)-(\intervalradius,0)$);
\coordinate (R) at ($(P)+(\intervalradius,0)$);

\coordinate (Pshift) at ($(P)+(\shiftamount,-\rowgap)$);
\coordinate (Lshift) at ($(L)+(\shiftamount,-\rowgap)$);
\coordinate (Rshift) at ($(R)+(\shiftamount,-\rowgap)$);

\draw[very thick] (L) -- (R);
\fill (L) circle (1.8pt);
\draw[fill=white,thick] (R) circle (1.8pt);
\draw ([yshift=-1mm]P) -- ([yshift=1mm]P);
\node[above=1.5mm] at (P) {$p$};
\node[endpoint label,left=1mm] at (L) {$p-\alpha$};
\node[endpoint label,right=1mm] at (R) {$p+\alpha$};

\coordinate (shift-start)
    at (\centerposition,{-\rowgap+\titleheight});
\coordinate (shift-end) at ($(shift-start)+(\shiftamount,0)$);
\draw[->,thick] ([yshift=3mm]shift-start)
    -- ([yshift=3mm]shift-end)
    node[midway,above] {shift by $u$};

\draw[very thick] (Lshift) -- (Rshift);
\fill (Lshift) circle (1.8pt);
\draw[fill=white,thick] (Rshift) circle (1.8pt);
\draw ([yshift=-1mm]Pshift) -- ([yshift=1mm]Pshift);
\node[above=1.5mm] at (Pshift) {$p+u$};
\node[endpoint label,left=1mm] at (Lshift) {$(p+u)-\alpha$};
\node[endpoint label,right=1mm] at (Rshift) {$(p+u)+\alpha$};


\end{tikzpicture}}
    \caption{$u$ is an $\alpha$-good shift for $p$}
    \label{fig:goodshift}
    \Description{A diagram illustrating how a shift affects intervals}
\end{figure}

\begin{definition}
Let $\alpha>0$. A shift $u\in[0,\varepsilon)$ is $\alpha$-good for $p\in[0,1]$ if the interval $[p-\alpha,p+\alpha)$ when shifted by $u$ is contained between some consecutive integer multiple of $\varepsilon$. Formally, there exists an integer $k\in\mathbb{Z}$ such that
\[[(p+u)-\alpha,(p+u)+\alpha)\subseteq[k\varepsilon,(k+1)\varepsilon)\]
\end{definition}

\begin{definition}
Let $u\in[0,\varepsilon)$. The $u$-rounding of $x\in\mathbb{R}$, denoted by $R_u(x)$, is defined as 
\[R_u(x)=\left\lfloor\frac{x+u}{\varepsilon}\right\rfloor\varepsilon\]
For a vector $\boldx\in\mathbb{R}^d$, the $u$-rounding of $\boldx$, denoted by $R_u(\boldx)$, is obtained by applying index-wise $u$-rounding to $\boldx$, i.e., if $\boldx=(x_1,\ldots,x_d)$ then $R_u(\boldx)=\left(R_u(x_1),\ldots,R_u(x_d)\right)$
\end{definition}
\begin{lemma}\label{lem:uroundingapproximation}
Let $u\in[0,\varepsilon)$. For any $x\in\mathbb{R}$,
\[|R_u(x)-x|<\varepsilon\]
For any $\boldx\in\mathbb{R}^d$,
\[\left\lVert R_u(\boldx)-\boldx\right\rVert_\infty<\varepsilon\]
\end{lemma}
\begin{proof}
For every $y\in\mathbb{R}$, we have $y-1<\lfloor y\rfloor\leq y$. Applying this with $y=\frac{x+u}{\varepsilon}$ and multiplying by $\varepsilon$ gives
\begin{align*}
x+u-\varepsilon&<R_u(x)\leq x+u\\
\shortintertext{
Since $u\in[0,\varepsilon)$, it holds that}
x-\varepsilon&<R_u(x)<x+\varepsilon
\end{align*}
and hence $|R_u(x)-x|<\varepsilon$. Applying this bound to each coordinate gives
\[\left\lVert R_u(\boldx)-\boldx\right\rVert_\infty=\max_{i\in[d]}|R_u(x_i)-x_i|<\varepsilon\qedhere\]
\end{proof}

Consider an interval $[p-\alpha,p+\alpha)$ of length $2\alpha\leq\varepsilon$. This interval need not be contained between some $[k\varepsilon,(k+1)\varepsilon)$, however by shifting it by $u$ that is $\alpha$-good, $[(p+u)-\alpha,(p+u)+\alpha)$ is now contained in some $[k\varepsilon,(k+1)\varepsilon)$. Crucially, all points in this shifted interval round down to $k\varepsilon$; this property is what we shall be using in our upper bound algorithms. We now formalize this and prove some additional properties that we will require for our algorithm.

\begin{lemma}\label{lem:goodshiftproperties}
Let $u\in[0,\varepsilon)$, $\alpha>0$ and $p,\hat p\in[0,1]$ such that $|\hat p-p|\leq\alpha$.
\begin{enumerate}
\item If $u$ is $\alpha$-good for $p$ then for any $x,y\in[p-\alpha,p+\alpha)$ it holds that $R_u(x)=R_u(y)$
\item If $u$ is $3\alpha$-good for $p$, then $u$ is $2\alpha$-good for $\hat p$.
\item If $u$ is $2\alpha$-good for $\hat p$, then $R_u(p)=R_u(\hat{p})$.
\end{enumerate}
\end{lemma}
\begin{proof}\leavevmode
\begin{enumerate}
\item It holds that $x+u$ and $y+u$ both belong to $[k\varepsilon,(k+1)\varepsilon)$ by definition of $u$ being $\alpha$-good for $p$, and therefore both shifted points lie in the same interval and round down to $k\varepsilon=R_u(x)=R_u(y)$.
\item Observe that 
\[
[(\hat p+u)-2\alpha,(\hat p+u)+2\alpha)
\subseteq[(p+u)-3\alpha,(p+u)+3\alpha)
\]
Thus assuming $u$ is $3\alpha$-good for $p$ implies that $u$ is $2\alpha$-good for $\hat p$ by definition.

\item Since $|p-\hat p|\leq\alpha<2\alpha$, both $p,\hat p\in[\hat p-2\alpha,\hat p+2\alpha)$. Because $u$ is $2\alpha$-good for $\hat p$, using item 1 with center $\hat p$ and radius $2\alpha$, gives $R_u(\hat p)=R_u(p)$.\qedhere
\end{enumerate}
\end{proof}

\begin{lemma}\label{lem:probabilityuniformshiftisnotalphagoodbound}
Let $\alpha\in\left(0,\frac{\varepsilon}{2}\right]$ and $p\in[0,1]$. If $U\sim\Unif([0,\varepsilon))$ then
\[\Pr_{U}(\textup{$U$ is not $\alpha$-good for $p$})=\frac{2\alpha}{\varepsilon}\]
Furthermore, the equation above also holds when $p$ is a random variable and is independent of $U$.
\end{lemma}
\begin{proof}
For simplicity assume $p=k\varepsilon$ for some $k\in\mathbb{N}$. Observe that a shift $u$ is $\alpha$-good if and only if $u\in[\alpha,\varepsilon-\alpha]$. Thus we have
\[\Pr_{U}(\textup{$U$ is not $\alpha$-good for $p$})=\Pr_U(U\notin[\alpha,\varepsilon-\alpha])=\frac{2\alpha}{\varepsilon}\]
In general, if $p$ is not an exact multiple of $\varepsilon$, then $u$ is $\alpha$-good if and only if $(p+u) \bmod\varepsilon\in[\alpha,\varepsilon-\alpha]$, we also have that $(p+U)\bmod \varepsilon\sim\Unif([0,\varepsilon))$, so the equality still holds.

For the furthermore part, averaging over all values of the random variable $p$ yields the result.
\begin{align*}
&\Pr_{U,p}(\textup{$U$ is not $\alpha$-good for $p$})\\
&\quad=\sum_x
\Pr_U(\textup{$U$ is not $\alpha$-good for $p$}\mid p=x)
\Pr_p(p=x)\\
&\quad=\frac{2\alpha}{\varepsilon}\sum_x\Pr_p(p=x)\\
&\quad=\frac{2\alpha}{\varepsilon}\tag*{\qedhere}
\end{align*}
\end{proof}

Suppose we have a single coin with bias $p\in[0,1]$. Getting an estimate $\hat p$ such that $\hat p\in[p-\alpha,p+\alpha)$ with high probability can be achieved by simply tossing the coin $\widetilde{\Theta}(\frac{1}{\alpha^2})$ times and outputting the empirical mean. We sample $U\sim\Unif([0,\varepsilon))$ and get iteratively better approximations $\alpha_j=\frac{\varepsilon}{2^{j+1}}$, till $U$ is $2\alpha_j$-good for $\hat p$, i.e., the interval $[\hat p-2\alpha_j,\hat p+2\alpha_j)$ when shifted by $U$ fits in some $[k\varepsilon,(k+1)\varepsilon)$, once this condition is met, and assuming our estimates are correct we have $R_U(\hat p)=R_U(p)$, i.e., the $u$-rounding $\hat p$ is exactly the $u$-rounding of $p$, the unknown bias. If this procedure takes too many samples, we can cut it off early and output $\bot$. Even though we don't know the bias $p$, we can with high probability get $R_U(p)$. 

The rounded value of $p$ crucially depends on $U$. Suppose $p=2.5\varepsilon$, if $u\in\left[0,\frac{\varepsilon}{2}\right)$ then $R_u(p)=2\varepsilon$, and if $u\in\left[\frac{\varepsilon}{2},\varepsilon\right)$ then $R_u(p)=3\varepsilon$. We now extend this idea to higher dimensions.
\begin{definition}\label{def:canonicallist}
Let $\boldp\in[0,1]^d$. The set $\calL_\boldp$ is defined as
\[\calL_\boldp=\{R_u(\boldp)\mid u\in[0,\varepsilon)\}\]
\end{definition}
\begin{lemma}\label{lem:canonicallistproperties}
For any $\boldp\in[0,1]^d$, the set $\calL_\boldp$ satisfies
\begin{enumerate}
\item $\forall\boldy\in\calL_\boldp\ldotp\left\lVert \boldy-\boldp\right\rVert_\infty<\varepsilon$
\item $|\calL_\boldp|\leq d+1$
\end{enumerate}
\end{lemma}
\begin{proof}\leavevmode
\begin{enumerate}
\item Let $\boldy\in\calL_\boldp$, then we have $\boldy=R_u(\boldp)$ for some $u\in[0,\varepsilon)$, by \Cref{lem:uroundingapproximation} we have
\[\left\lVert \boldy-\boldp\right\rVert_\infty=\left\lVert R_u(\boldp)-\boldp\right\rVert_\infty<\varepsilon\]
\item As $u$ increases over $[0,\varepsilon)$, each coordinate $R_u(p_i)$ changes at most once. Thus the vector changes at most $d$ times and takes at most $d+1$ values.\qedhere
\end{enumerate}
\end{proof}

\paragraph{Example.} Consider $d=4$ and $\boldp=(\frac{\varepsilon}{5},\frac{2\varepsilon}{5},\frac{3\varepsilon}{5},\frac{4\varepsilon}{5})$ then
\begin{enumerate}
\item If $u\in[0,\frac{\varepsilon}{5})$ then $R_u(\boldp)=(0,0,0,0)$
\item If $u\in[\frac{\varepsilon}{5},\frac{2\varepsilon}{5})$ then $R_u(\boldp)=(0,0,0,\varepsilon)$
\item If $u\in[\frac{2\varepsilon}{5},\frac{3\varepsilon}{5})$ then $R_u(\boldp)=(0,0,\varepsilon,\varepsilon)$
\item If $u\in[\frac{3\varepsilon}{5},\frac{4\varepsilon}{5})$ then $R_u(\boldp)=(0,\varepsilon,\varepsilon,\varepsilon)$
\item If $u\in[\frac{4\varepsilon}{5},\varepsilon)$ then $R_u(\boldp)=(\varepsilon,\varepsilon,\varepsilon,\varepsilon)$
\end{enumerate}
Thus, in this example $|\calL_\boldp|=d+1$ exactly.

Our algorithms will attempt to output some element in the set $\calL_\boldp$ with high probability. Let $Y$ be a random variable denoting the output of a list replicable algorithm which belongs to $\mathbb{R}^d\cup\{\bot\}$, where $\bot$ is a failure symbol. The output can be categorized into 3 mutually exclusive cases
\begin{enumerate}
\item $Y\in \calL_\boldp$, this is the desired event
\item $Y=\bot$, not a valid member of $\mathbb{R}^d$
\item $Y\notin \calL_\boldp\cup\{\bot\}$, a valid member $\mathbb{R}^d$ but not in the list
\end{enumerate}
For the purpose of analysis, we will be upper bounding the quantities $\Pr(Y=\bot)$ and $\Pr(Y\notin \calL_\boldp\cup\{\bot\})$ as it is easier to argue over. However, it might be easier to interpret the bounds as follows: suppose
\begin{align*}
\Pr(Y=\bot)                             &\leq\frac{1}{4}\\
\Pr(Y\notin \calL_\boldp\cup\{\bot\})   &\leq\eta
\end{align*}
This means that, with probability at least $3/4$ the algorithm will return a vector in $\mathbb{R}^d$, and conditioned on witnessing that $Y\neq\bot$, the probability that $Y\in\calL_\boldp$ is at least $1-\frac{4\eta}{3}$. Formally,
\[\Pr(Y\notin\calL_\boldp\mid Y\neq\bot)=\frac{\Pr(Y\notin\calL_\boldp\cup \{\bot\})}{\Pr(Y\neq\bot)}\leq\frac{4\eta}{3}\]

\subsection{\texorpdfstring{Upper Bound for $(d+1)$-List Replicable Algorithms}{Upper Bound for (d+1)-List Replicable Algorithms}}\label{subsec:upperboundford1listreplicablealgorithms}

We prove the following theorem
\begin{theorem}\label{thm:upperboundmaintheorem}
Let $d\in\mathbb{N}$ and $\varepsilon,\delta\in(0,1)$. There exists a $(d+1,\varepsilon,\delta)$-list replicable algorithm for the $d$-$\CoinBiasEstimationProblem$ that uses $N=O\left(\frac{d^2}{\varepsilon^2}\log\frac{d}{\delta}\log\frac{1}{\delta}\right)$ samples.
\end{theorem}
We first prove the existence of an algorithm whose expected sample complexity is $\EE[N]=\widetilde{O}\left(\frac{d^2}{\varepsilon^2}\right)$. Then using standard boosting techniques, we get \Cref{thm:upperboundmaintheorem}.

Define the following parameters
\begin{alignat*}{2}
\varepsilon &\in\left(0,1\right)\\
\eta        &\in\left(0,1/8\right)\\
m           &=\lceil\log_2(24d)\rceil\\
\alpha_j    &=\frac{\varepsilon}{2^{j+1}}&&\quad\textup{for $j\in[m]$}\\
n_j         &=\left\lceil\frac{1}{2\alpha_j^2}\ln\frac{2dm}{\eta}\right\rceil&&\quad\textup{for $j\in[m]$}\\
M           &=\frac{1536d^2}{\varepsilon^2}\ln\frac{2dm}{\eta}=O\left(\frac{d^2}{\varepsilon^2}\ln\frac{d}{\eta}\right)
\end{alignat*}
We now describe a $(d+1,\varepsilon,O(1))$-list replicable algorithm for the $d$-$\CoinBiasEstimationProblem$ whose expected sample complexity is bounded above $M=O\left(\frac{d^2}{\varepsilon^2}\ln\frac{d}{\eta}\right)$. We use the same $\varepsilon$ to perform rounding as described \Cref{subsec:shiftsroundingandcanonicallist}.

\begin{algorithm}[ht]
\caption{$(d+1,\varepsilon,O(1))$-List Replicable Algorithm with $\widetilde{O}\left(\frac{d^2}{\varepsilon^2}\right)$ expected total sample complexity}
\label{alg:expectedsamplecomplexityalgorithm}
Draw $U\sim\operatorname{Unif}([0,\varepsilon))$\;

$(y_1,\ldots,y_d)\gets(\bot,\ldots,\bot)$

$m=\lceil\log_2(24d)\rceil$

$A\gets[d]$

\For{$j=1,\ldots,m$}{
    \ForEach{$i\in A$}{
        $\alpha_j\gets\frac{\varepsilon}{2^{j+1}}$
    
        Toss coin $i$ independently $n_j=\left\lceil\frac{1}{2\alpha_j^2}\ln\frac{2dm}{\eta}\right\rceil$ times and let $\hat p_i^{(j)}$ be the empirical mean

        \If{$U$ is $2\alpha_j$-good for $\hat p_i^{(j)}$}{
            $y_i\gets R_U(\hat p_i^{(j)})$
            
            $A\gets A\setminus\{i\}$
        }
    }
    \If{$A=\varnothing$}{
        \Return $(y_1,\ldots,y_d)$
    }
}
\Return $\bot$
\end{algorithm}

\begin{theorem}
\Cref{alg:expectedsamplecomplexityalgorithm} has the following properties
\begin{align*}
\Pr(Y=\bot)&\leq\frac{1}{4}\\
\Pr(Y\notin \calL_\boldp\cup\{\bot\})&\leq\eta
\end{align*}
The expected sample complexity of the algorithm is bounded above by
\begin{align*}
\EE[N]&\leq M=O\left(\frac{d^2}{\varepsilon^2}\log\frac{d}{\eta}\right)
\end{align*}
\end{theorem}

Suppose \Cref{alg:expectedsamplecomplexityalgorithm} has sample access to $d$ coins with bias vector $\boldp=(p_1,\ldots,p_d)$. Further, let $N_i$ denote the number of tosses of coin $i$ and let $N=\sum_{i=1}^d N_i$ denote the total number of tosses. Let $Y$ be the random variable denoting the output of the algorithm.

For each coin $i\in[d]$, we will get progressively smaller estimates $\alpha_j$ as the number of rounds $j$ increases. We will get at most $dm$ estimates across the entire algorithm's execution, $m$ estimates for each of the $d$ coins. So we picked $n_j=\left\lceil\frac{1}{2\alpha_j^2}\ln\frac{2dm}{\eta}\right\rceil$ so that for the $i$th coin in $j$th round the estimate $\hat p_{i}^{(j)}$ is $\alpha_j$ close to $p_i$ with probability at least $1-\frac{\eta}{dm}$. Union bounding across all coins and rounds: every $|\hat p_{i}^{(j)}-p_i|\leq\alpha_j$ happens with probability at least $1-\eta$.

For simplicity of analysis, we imagine that an independent block of $n_j$ tosses is assigned to every $i\in[d]$ and $j\in[m]$, independently of $U$, whether or not the algorithm uses it. Let $\hat p_{i}^{(j)}$ denote the empirical mean of each block, whether or not this was actually used. Only blocks actually used by the algorithm count toward its sample complexity. Formally define the following events
\begin{alignat*}{2}
G_{i}&=\{\textup{$U$ is $3\alpha_m$-good for $p_i$}\}&&\qquad\textup{for $i\in[d]$}\\
H_{i,j}&=\{\hat p_{i}^{(j)}\in[p_i-\alpha_j,p_i+\alpha_j]\}&&\qquad\textup{for $i\in[d],j\in[m]$}\\
G&=\cap_{i}G_{i}\\
H&=\cap_{i,j}H_{i,j}
\end{alignat*}
i.e., $G$ is the event that $U$ is $3\alpha_m$-good $p_i$ for all $i$ and $H$ is the event that $\hat p_{i}^{(j)}$ is $\alpha_j$-close to $p_i$ for all $i,j$.

\begin{lemma}\leavevmode
\begin{enumerate}
\item $\Pr(G^c)\leq\frac{1}{8}$
\item $\Pr(H^c)\leq\eta$
\item $\Pr(G^c\cup H^c)\leq\frac{1}{8}+\eta\leq\frac{1}{4}$
\end{enumerate}
\end{lemma}
\begin{proof}
\leavevmode
\begin{enumerate}
\item By union bound and \Cref{lem:probabilityuniformshiftisnotalphagoodbound} on $G_i^c=\{\textup{$U$ is not $3\alpha_m$-good for $p_i$}\}$
\[
\Pr(G^c)
\leq\sum_{i=1}^d\Pr(G_i^c)
=d\cdot\frac{2\cdot(3\alpha_m)}{\varepsilon}
=\frac{3d}{2^m}
\leq\frac18.
\]
The last inequality is obtained by using $m=\lceil\log_2(24d)\rceil\geq \log_2(24d)$.
\item By Chernoff Bound, for any $i,j$
\[\Pr(H_{i,j}^c)=\Pr(|\hat p_{i}^{(j)}-p_i|>\alpha_j)\leq 2e^{-2n_j\alpha_j^2}\leq\frac{\eta}{dm}\]
By union bound, we have
\[\Pr(H^c)\leq\sum_{i,j}\Pr(H_{i,j}^c)\leq dm\cdot\frac{\eta}{dm}=\eta\]
\item By union bound, we have
\[\Pr(G^c\cup H^c)\leq\frac{1}{8}+\eta\leq\frac{1}{4}\]
The last inequality is because we assumed $\eta\leq \frac{1}{8}$\qedhere
\end{enumerate}
\end{proof}

For a coin $i\in[d]$, let $Z_i$ be the random variable denoting first round $j$ for which $U$ is $2\alpha_j$-good for $\hat p_i^{(j)}$, if it never happens we let $Z_i=\infty$. Observe that $Z_i\in\{1,\ldots,m\}\cup\{\infty\}$. We say coin $i$ is successful if $Z_i\neq\infty$. Furthermore, for all rounds $j\leq Z_i$ we say that coin $i$ was active, meaning that we had to draw $n_j$ samples for coin $i$ for that round. Note that the set $A$ consists of indices of those coins that are active. If coin $i$ is successful then $y_{i}=R_U(\hat p_{i}^{(Z_i)})$. If some coin $i$ was not successful, then $y_i=\bot$ and consequently $Y=\bot$.

\begin{lemma}\leavevmode
\begin{enumerate}
\item On $H$, it holds that $Y\in\{R_U(\boldp),\bot\}$. In particular,
\[\Pr(Y\notin \calL_{\boldp}\cup\{\bot\})\leq\eta\]
\item On $G\cap H$, it holds that $Y=R_U(\boldp)$. In particular,
\[\Pr(Y=\bot)\leq\frac{1}{4}\]
\end{enumerate}
\end{lemma}
\begin{proof}\leavevmode
\begin{enumerate}
\item First note that unconditionally, for every coin $i\in[d]$, $y_{i}\in\{R_U(\hat p_{i}^{(1)}),\ldots,R_U(\hat p_{i}^{(m)}),\bot\}$. The exact value of $y_i$ depends on $Z_i$: if $Z_i\neq\infty$ then $y_i=R_U(\hat p_{i}^{(Z_i)})$. Even if a single coin is unsuccessful, we trivially have $Y=\bot$. Thus we assume all coins are successful and we show that $Y=R_{U}(\boldp)$, i.e., we show that for all $i$, it holds that $y_i=R_U(p_i)$. Using this, we have that for all $i$, at round $Z_i$, it holds that $U$ is $2\alpha_{Z_i}$-good for $\hat p_i^{(Z_i)}$ and $y_i=R_U(\hat p_{i}^{(Z_i)})$ by construction of our algorithm. 

Condition on $H$ and fix a coin $i$; we have that $|\hat p_{i}^{(j)}-p_i|\leq\alpha_j$ for all $j\in[m]$. In particular in at round $Z_i$, we have $|\hat p_{i}^{(Z_i)}-p_i|\leq\alpha_{Z_i}$. Using item 3 of \Cref{lem:goodshiftproperties}, we have that $y_i=R_U(\hat p_{i}^{(Z_i)})=R_U(p_i)$. Since this argument holds for all coins, we have $Y=R_U(\boldp)$. If $Y$ is non-$\bot$ output and was not in $\calL_p$, then some estimate $\hat p_i^{(j)}$ was not $\alpha_j$-close to $p_i$, thus we have
\[\Pr\left(Y\notin \calL_{\boldp}\cup\{\bot\}\right)\leq\Pr(H^c)\leq\eta\]

\item Conditioning on $H$, yielded $Y\in\{R_U(\boldp),\bot\}$. We show that on $G\cap H$, it holds that $Y\neq \bot$, thus yielding that $Y=R_U(\boldp)$. Fix a coin $i\in[d]$ and consider the last round $m$, in this round, because of conditioning on $G$ we have that $U$ is $3\alpha_m$-good for $p_i$, and because of conditioning on $H$ we have that $|\hat p_{i}^{(m)}-p_i|\leq \alpha_m$. Using item 2 of \Cref{lem:goodshiftproperties} we have that $U$ is also $2\alpha_m$-good for $\hat p_{i}^{(m)}$, which is exactly the definition of being successful at round $m$, thus if coin $i$ wasn't successful by round $m-1$ on $G\cap H$, on round $m$ it becomes successful, i.e., $Z_i\leq m$, thus $y_i\neq \bot$. Since this argument holds for all coins, we have that $Y\neq\bot$ and it holds that $Y=R_U(\boldp)\in \calL_{\boldp}$. Thus we have
\[\Pr(Y\in \calL_\boldp)\geq\Pr(Y=R_U(\boldp))\geq \Pr(G\cap H)\geq\frac{3}{4}\]
Finally 
\[\Pr(Y=\bot)\leq\Pr(Y\notin \calL_\boldp)\leq\frac{1}{4}\qedhere\]
\end{enumerate}
\end{proof}

\begin{lemma}\label{lem:expectedtotalsamplecomplexityalgorithmsamplexomplexitybound}
The expected sample complexity is bounded by
\[\EE[N]\leq M=\frac{1536d^2}{\varepsilon^2}\ln\frac{2dm}{\eta}=O\left(\frac{d^2}{\varepsilon^2}\log\frac{d}{\eta}\right)\]
\end{lemma}
\begin{proof}
Fix a coin $i\in[d]$. 
If coin $i$ is active in round $j\geq2$, then it was active in round $j-1$ and failed the test in that round, since $\hat p_{i}^{(j-1)}$ is independent of $U$ we have
\begin{align*}
&\Pr(\textup{coin $i$ was active at round $j$})\\
&\quad\leq\Pr(\textup{$U$ is not $2\alpha_{j-1}$-good for $\hat p_i^{(j-1)}$})\\
&\quad=\frac{2\cdot 2\alpha_{j-1}}{\varepsilon}\tag{\Cref{lem:probabilityuniformshiftisnotalphagoodbound}}\\
&\quad=\frac{4}{2^{j}}\tag{$\alpha_{j-1}=\frac{\varepsilon}{2^{j}}$}
\end{align*}
The upper bound also trivially holds when $j=1$. Applying it to the following, we have
\begin{align*}
\EE[N_i]&=\sum_{j=1}^{m}n_j\Pr(\textup{coin $i$ was active at round $j$})\\
        &\leq\sum_{j=1}^{m}\left\lceil\frac{1}{2\alpha_j^2}\ln\frac{2dm}{\eta}\right\rceil\cdot\frac{4}{2^j}\\
        &\leq\sum_{j=1}^{m}\left(\frac{1}{\alpha_j^2}\ln\frac{2dm}{\eta}\right)\cdot\frac{4}{2^j}\tag{$x\geq1\implies\lceil x\rceil\leq2x$}\\
        &=\frac{16}{\varepsilon^2}\ln\frac{2dm}{\eta}\sum_{j=1}^{m}2^j\tag{$\alpha_j=\frac{\varepsilon}{2^{j+1}}$}\\
        &\leq \frac{32\cdot2^m}{\varepsilon^2}\ln\frac{2dm}{\eta}\tag{$\sum_{j=1}^{m}2^j\leq 2^{m+1}$}\\
        &\leq \frac{1536d}{\varepsilon^2}\ln\frac{2dm}{\eta}
\end{align*}
The last inequality uses $m=\lceil\log_2(24d)\rceil\leq \log_2(24d)+1$. Finally, using linearity of expectation across all coins, we have
\begin{align*}
\EE[N]  &=\sum_{i=1}^{d}\EE[N_i]\\
        &\leq \sum_{i=1}^{d}\frac{1536d}{\varepsilon^2}\ln\frac{2dm}{\eta}\\
        &=\frac{1536d^2}{\varepsilon^2}\ln\frac{2dm}{\eta}\\
        &=M=O\left(\frac{d^2}{\varepsilon^2}\log\frac{d}{\eta}\right)
\end{align*}
The asymptotic upper bound is obtained by observing that $m=O(\log d)\subseteq O(d)$ and therefore $\ln\left(\frac{2dm}{\eta}\right)=O\left(\log\frac{d}{\eta}\right)$.
\end{proof}

We now conclude the \Cref{thm:upperboundmaintheorem}. We have an algorithm that is $(d+1,\varepsilon,O(1))$-list replicable for $d$-$\CoinBiasEstimationProblem$ with expected sample complexity $\EE[N]\leq M=\widetilde{O}\left(\frac{d^2}{\varepsilon^2}\right)$ satisfying $\Pr(Y=\bot)\leq\frac{1}{4}$ and $\Pr(Y\notin\calL_\boldp\mid Y\neq\bot)=O(\eta)$. We can convert the expected sample complexity algorithm into an algorithm with fixed sample complexity by using standard techniques. Capping the sample complexity to $B:=\left\lceil4M\right\rceil$ increases the probability we see $\bot$ by at most $\frac{1}{4}$ by Markov's Inequality. Thus we repeat this capped algorithm $T=\left\lceil\log_2\frac{2}{\delta}\right\rceil$ times and return the first non-$\bot$ output. We see a non-$\bot$ output with probability at least $1-\frac{\delta}{2}$, and pick $\eta=\frac{\delta}{4}$ so that conditioned on seeing a non-$\bot$ output, the probability that it belongs to $\calL_\boldp$ is at least $1-\frac{\delta}{2}$. A more detailed analysis can be found in \Cref{sec:upperboundwithfixedsamplecomplexity}.

\subsection{\texorpdfstring{Upper Bound for $k$-List Replicable Algorithms}{Upper Bound for k-List Replicable Algorithms}}\label{subsec:upperboundforklistreplicablealgorithms}
\begin{theorem}\label{thm:upperboundmaintheoremklist}
Let $d,k\in\mathbb{N}$ such that $k\in[d+1,2^d]$ and $\varepsilon,\delta\in(0,1)$. There exists a $(k,\varepsilon,\delta)$-list replicable algorithm for the $d$-$\CoinBiasEstimationProblem$ that uses $N=O\left(\frac{d^2\log d}{\varepsilon^2\log k}\log^2\frac{d}{\delta}\right)$ samples.
\end{theorem}

Let $\varepsilon\in(0,1)$, $\delta\in(0,\frac{1}{2})$ and let $\ell\in\mathbb{N}$ be decided later. Consider the standard partitioning $\calP$ of $[d]$ into $\ell$ consecutive blocks $B_1,\ldots, B_\ell$ each of size $d_i:=|B_i|\leq\left\lceil \frac{d}{\ell}\right\rceil$. Run algorithm from \Cref{thm:upperboundmaintheorem} independently for each block of $d_i$ coins with parameters $\left(d_i+1,\varepsilon,\frac{\delta}{\ell}\right)$. If any block returns $\bot$, we return $\bot$. Otherwise, we return the concatenation of the outputs of each block.

Let $Y_1,\ldots, Y_\ell$ be the outputs of each of the blocks. Also, let $Y$ and $N$ be the random variables denoting the output of this algorithm and the sample complexity. 

Fix $\boldp\in[0,1]^d$ and for $i\in[\ell]$, let $\boldp^{(i)}=(p_j)_{j\in B_i}\in[0,1]^{d_i}$ denote the bias vector in $B_i$. The list achieved by this algorithm is
\[\calL_{\boldp,\calP}=\calL_{\boldp^{(1)}}\times\ldots\times\calL_{\boldp^{(\ell)}}\]
where $\calL_{\boldp^{(i)}}$ is the canonical list as defined in \Cref{def:canonicallist}. We identify each element of this Cartesian product with the concatenation of its block vectors in $\mathbb{R}^d$.

Fix $i\in[\ell]$, by \Cref{lem:canonicallistproperties} each element of $\calL_{\boldp^{(i)}}$ is still $\varepsilon$-close to $\boldp^{(i)}$ and the size of this set is $|\calL_{\boldp^{(i)}}|\leq d_i+1\leq \left\lceil\frac{d}{\ell}\right\rceil+1$. Therefore, for each $\boldy\in\calL_{\boldp,\calP}$, it holds that $\lVert\boldy-\boldp\rVert_\infty<\varepsilon$ and the size of $\calL_{\boldp,\calP}$ is bounded above by
\begin{equation}
|\calL_{\boldp,\calP}|\leq\left(\left\lceil\frac{d}{\ell}\right\rceil+1\right)^{\ell}\label{eqn:listsize}
\end{equation}

The sample complexity is bounded by
\begin{align}
N&=\sum_{i=1}^{\ell}O\left(\frac{|B_i|^2}{\varepsilon^2}\log\frac{|B_i|\ell}{\delta}\log\frac{\ell}{\delta}\right)\nonumber\\
    &=\sum_{i=1}^{\ell}O\left(\frac{d^2}{\ell^2\varepsilon^2}\log\frac{d}{\delta}\log\frac{\ell}{\delta}\right)\nonumber\\
    &=O\left(\frac{d^2}{\ell\varepsilon^2}\log\frac{d}{\delta}\log\frac{\ell}{\delta}\right)\label{eqn:samplecomplexity}
\end{align}

By \Cref{thm:upperboundmaintheorem}, we have $\forall i\in[\ell]\ldotp \Pr(Y_i\in\calL_{\boldp^{(i)}})\geq1-\frac{\delta}{\ell}$. Taking a union bound over all $\ell$ blocks gives us $\Pr(\forall i\in[\ell]\ldotp Y_i\in \calL_{\boldp^{(i)}})\geq1-\delta$. In particular, when $\forall i\in[\ell]\ldotp Y_i\in \calL_{\boldp^{(i)}}$ then $Y\in\calL_{\boldp,\calP}$. Therefore
\[\Pr(Y\in\calL_{\boldp,\calP})\geq1-\delta\]

Thus we pick $\ell:=\left\lfloor\log_{d+1}k\right\rfloor$ so that the list size in \Cref{eqn:listsize} simplifies to
\begin{align*}
|\calL_{\boldp,\calP}|  &\leq\left(\left\lceil\frac{d}{\ell}\right\rceil+1\right)^{\ell}\\
                        &\leq\left(d+1\right)^{\ell}\tag{$\left\lceil\frac{d}{\ell}\right\rceil\leq d$}\\
                        &\leq\left(d+1\right)^{\log_{d+1}k}\\
                        &\leq k
\end{align*}

Using $\ell=\left\lfloor\log_{d+1}k\right\rfloor\geq\frac{1}{2}\log_{d+1}k$, the sample complexity in \Cref{eqn:samplecomplexity} simplifies to
\begin{align*}
N   &=O\left(\frac{d^2}{\ell\varepsilon^2}\log\frac{d}{\delta}\log\frac{\ell}{\delta}\right)\\
    &=O\left(\frac{d^2\log(d+1)}{\varepsilon^2\log k}\log\frac{d}{\delta}\log\frac{\frac{\log k}{\log(d+1)}}{\delta}\right)\\
    &=O\left(\frac{d^2\log d}{\varepsilon^2\log k}\log^2\frac{d}{\delta}\right)\\
    &=\widetilde{O}\left(\frac{d^2}{\varepsilon^2\log k}\right)
\end{align*}
In the second last step, we assumed $k\leq 2^d$, thus $\frac{\log k}{\log (d+1)}\leq d$.

This concludes the proof of \Cref{thm:upperboundmaintheoremklist}.

\section{Conclusion}
Our results establish that the sample complexity of $(d+1,\varepsilon,\delta)$-list replicable algorithms for the $d$-$\CoinBiasEstimationProblem$ is $\widetilde{\Theta}\left(\frac{d^2}{\varepsilon^2}\right)$. More generally, we establish a sample complexity bound of $\widetilde{\Theta}\left(\frac{d^2}{\varepsilon^2\log k}\right)$ for $(k,\varepsilon,\delta)$-list replicable algorithms for the $d$-$\CoinBiasEstimationProblem$ where $k\in[d+1,2^{d}]$. Our upper bounds improve the polynomial dependence on $d$ from cubic to quadratic compared with \cite{DPVV2023, WDPRV2024}. More generally, our upper and lower bounds determine the sample complexity throughout $k\in[d+1,2^d]$ up to polylogarithmic factors.

\paragraph{Open Problems.}\Cref{sec:upperbound} exhibits algorithms whose sample complexity for each coin is adaptive in the sense that the sample complexity of each $N_i$ is not fixed in advance. So the question of whether there exists a non-adaptive algorithm that is $(d+1,\varepsilon,\delta)$-list replicable for the $d$-$\CoinBiasEstimationProblem$ with sample complexity $\widetilde{O}\left(\frac{d^2}{\varepsilon^2}\right)$ remains open.

\paragraph{Acknowledgements.}
Vinod is supported in part by NSF awards 2342244, 2413848, and a UNL Grand Challenges Grant. Pavan, Michael, and Atieh are supported by NSF awards 2342245 and 2413849.

\paragraph{AI methodology.} We used ChatGPT-6 Astra and Google Gemini 3.6. Specifically, the LLMs helped us identify some of the bounds in \Cref{subsec:changeofmeasureforcointosses,sec:generallowerbound}, and ideas from LLMs helped us to arrive at the upper results in \Cref{sec:upperbound}. 
The authors wrote the manuscript with limited assistance from LLMs. We verified the correctness and originality of all content, including the accuracy of the references.


\bibliographystyle{alpha}
\bibliography{bibliography}

\appendix
\crefalias{section}{appendix}
\crefname{appendix}{appendix}{appendices}
\Crefname{appendix}{Appendix}{Appendices}
\section{Upper Bound with Fixed Sample Complexity}
\label{sec:upperboundwithfixedsamplecomplexity}

\subsection{Capping the Sample Complexity}\label{subsec:cappingthesamplecomplexity}
Let $\varepsilon,\eta$ be the same parameters as in \Cref{alg:expectedsamplecomplexityalgorithm}. Define the parameters
\begin{align*}
M           &=\frac{1536d^2}{\varepsilon^2}\ln\frac{2dm}{\eta}\\
B           &=\lceil 4M\rceil
\end{align*}
where $m=\lceil\log_2(24d)\rceil$ and $M$ is the same upper bound on the expected number of samples from \Cref{alg:expectedsamplecomplexityalgorithm}.

\begin{algorithm}[ht]
\caption{$(d+1,\varepsilon,O(1))$-List Replicable Algorithm with sample complexity $N=\widetilde{O}\left(\frac{d^2}{\varepsilon^2}\right)$}
\label{alg:cappedsamplecomplexityalgorithm}
Run \Cref{alg:expectedsamplecomplexityalgorithm} with a cap of $B=\lceil4M\rceil$ samples\;

\eIf{\Cref{alg:expectedsamplecomplexityalgorithm} needs more than $B$ samples}{
    \Return $\bot$
}{
    \Return Output of the \Cref{alg:expectedsamplecomplexityalgorithm}
}
\end{algorithm}
\begin{theorem}
\Cref{alg:cappedsamplecomplexityalgorithm} has the following properties
\begin{align*}
\Pr(Y=\bot)&\leq\frac{1}{2}\\
\Pr(Y\notin \calL_\boldp\cup\{\bot\})&\leq\eta
\end{align*}
The sample complexity of the algorithm is bounded above by
\[N\leq B=\lceil4M\rceil=O\left(\frac{d^2}{\varepsilon^2}\log\frac{d}{\eta}\right)\]
\end{theorem}
For \Cref{alg:cappedsamplecomplexityalgorithm}, let $N$ be the random variable denoting the total number of samples and $Y$ be the random variable denoting the output.

\begin{proof}
Imagine running both algorithms with the same shift and samples. Their outputs agree whenever the uncapped algorithm uses at most $B$ samples. By \Cref{lem:expectedtotalsamplecomplexityalgorithmsamplexomplexitybound} and applying Markov's Inequality, we get
\[\Pr(\textup{\Cref{alg:expectedsamplecomplexityalgorithm} requires more than $B$ samples})\leq\frac{1}{4}\]
The probability that \Cref{alg:expectedsamplecomplexityalgorithm} outputs $\bot$ is at most $\frac{1}{4}$, the probability $Y=\bot$ increases by at most $\frac{1}{4}$ when the algorithm requires more than $B$ samples. Thus we have
\[\Pr(Y=\bot)\leq\frac{1}{4}+\frac{1}{4}\leq\frac{1}{2}\]
Let $N$ be the sample complexity of \Cref{alg:cappedsamplecomplexityalgorithm}. We cap the algorithm's sample complexity bound at $\lceil 4M\rceil$, so trivially $N\leq \lceil 4M\rceil$ always holds. This cap can only change outputs from $\mathbb{R}^d$ to $\bot$, but not the other way around. So the following still holds
\[\Pr(Y\notin \calL_\boldp\cup\{\bot\})\leq\eta\qedhere\]
\end{proof}



\subsection{Boosting the Success Probability}

Define the parameters
\begin{align*}
\varepsilon &\in(0,1)\\
\delta      &\in(0,1/2)\\
T           &=\left\lceil\log_2\frac{2}{\delta}\right\rceil\\
\eta        &=\frac{\delta}{4}
\end{align*}
So that $\eta\in(0,1/8]$. A single execution of \Cref{alg:cappedsamplecomplexityalgorithm} may return $\bot$, so we run the same algorithm independently up to $T$ times till it produces an element of $\mathbb{R}^d$. We return the first non-$\bot$ output, and return $\bot$ only if all $T$ attempts do so.

\begin{algorithm}[ht]
\caption{$(d+1,\varepsilon,\delta)$-List Replicable Algorithm with sample complexity $N=\widetilde{O}\left(\frac{d^2}{\varepsilon^2}\right)$}
\label{alg:finallistreplicablealgorithm}
\For{$t=1,\ldots, T$} {
    Run an i.i.d. instance of \Cref{alg:cappedsamplecomplexityalgorithm} with $\varepsilon$, $\eta=\frac{\delta}{4}$

    Let $Y_t$ be its output
    
    \If{$Y_t\neq \bot$}{
        \Return $Y_t$
    }
}
\Return $\bot$
\end{algorithm}
\Cref{alg:finallistreplicablealgorithm} outputs $\bot$ only if all $T$ instances output $\bot$, which happens with probability at most $2^{-T}\leq\delta/2$. Each instance returns a valid member of $\mathbb{R}^d$ with probability at least $1/2$ and returns a vector outside $\calL_\boldp$ with probability at most $\eta$. Thus, among its non-$\bot$ outputs, the probability of being outside the list is at most $2\eta=\delta/2$. Therefore, the probability of returning $\bot$ or a vector outside the list is at most $\delta$. We now prove the correctness and sample complexity bound of this algorithm formally.
\begin{theorem}
\Cref{alg:finallistreplicablealgorithm} is $(d+1,\varepsilon,\delta)$-list replicable with list $\calL_\boldp$, i.e.,
\[\Pr(Y\in \calL_\boldp)\geq1-\delta\]
where $\calL_\boldp=\{R_u(\boldp)\mid u\in[0,\varepsilon)\}$ satisfies $|\calL_\boldp|\leq d+1$ and each $\boldy\in\calL_\boldp$ satisfies $\left\lVert \boldy-\boldp\right\rVert_\infty<\varepsilon$.

The sample complexity of the algorithm is bounded above by
\[N\leq BT=O\left(\frac{d^2}{\varepsilon^2}\log\frac{d}{\delta}\log\frac{1}{\delta}\right)\]
\end{theorem}
\begin{proof}
The two properties of $\calL_\boldp$ follow directly from \Cref{lem:canonicallistproperties}. We prove the remaining properties. For simplicity of analysis, we imagine generating all $T$ independent instances $Y_1,\ldots,Y_T$, the algorithm returns the first non-$\bot$ output. 

Let $K$ be the random variable denoting the first index $t$ for which $Y_t\neq\bot$; if it doesn't exist, then $K=\infty$. By using properties of \Cref{alg:cappedsamplecomplexityalgorithm}, note that for any $t\in[T]$
\begin{align*}
\Pr(Y\notin \calL_\boldp\mid K=t)&=\Pr(Y_t\notin\calL_\boldp\mid Y_t\neq\bot)\\
&=\frac{\Pr(Y_t\notin\calL_\boldp\cup\{\bot\})}{\Pr(Y_t\neq\bot)}\\
&\leq2\Pr(Y_t\notin\calL_\boldp\cup\{\bot\})\tag{$\Pr(Y_t\neq\bot)\geq\frac{1}{2}$}\\
&\leq2\eta\tag{$\Pr(Y_t\notin\calL_\boldp\cup\{\bot\})\leq\eta$}\\
&\leq\frac{\delta}{2}
\end{align*}
And by independence we have
\begin{align*}
\Pr(K=\infty)   &=\Pr(Y_1=\bot,\ldots,Y_T=\bot)\\
                &=\prod_{t=1}^{T}\Pr(Y_t=\bot)\\
                &\leq\left(\frac{1}{2}\right)^T\\
                &\leq\frac{\delta}{2}
\end{align*}
And finally
\begin{align*}
\Pr(Y\notin\calL_\boldp)&=\Pr(Y\notin\calL_\boldp,K=\infty)+\sum_{t=1}^{T}\Pr(Y\notin\calL_\boldp,K=t)\\
    &=\Pr(K=\infty)+\sum_{t=1}^{T}\Pr(Y\notin\calL_\boldp\mid K=t)\Pr(K=t)\\
    &\leq\Pr(K=\infty)+\frac{\delta}{2}\sum_{t=1}^{T}\Pr(K=t)\\
    &\leq\frac{\delta}{2}+\frac{\delta}{2}\cdot1\\
    &\leq\delta
\end{align*}
Each attempt uses at most $B=\lceil4M\rceil$ tosses, and there are at most $T$ attempts. Substituting $\eta=\frac{\delta}{4}$ in the expression of $B$, the total sample count is bounded by
\begin{align*}
N   &\leq BT\\
    &= O\left(\frac{d^2}{\varepsilon^2}\log\frac{4d}{\delta}\right)\left\lceil\log_2\frac{2}{\delta}\right\rceil\\
    &= O\left(\frac{d^2}{\varepsilon^2}\log\frac{d}{\delta}\log\frac{1}{\delta}\right)\tag*{\qedhere}
\end{align*}
\end{proof}


\section{General Lower Bound}\label{sec:generallowerbound}
In this section, we prove the lower bound result for arbitrary algorithms.
\begin{theorem}[\Cref{thm:lowerboundmaintheorem} Restated]
Let $\varepsilon,\delta\in(0,1/8)$. For sufficiently large $d\in\mathbb{N}$, every $(d+1,\varepsilon,\delta)$-list replicable algorithm for the $d$-$\CoinBiasEstimationProblem$ requires $N=\Omega\left(\frac{d^2}{\varepsilon^2\log d}\right)$ samples.
\end{theorem}

Let $\calA$ be a randomized sampling-based algorithm that draws $N$ samples from $d$ coins with unknown bias and is $(d+1,\varepsilon,\delta)$-list replicable. The algorithm may choose which coin to toss based on its internal randomness and previous outcomes. Let $Y$ be the random variable denoting the algorithm's output, which belongs to the set $\calY$. Let $W$ be the random variable denoting the algorithm's internal randomness. Let $H_0=W$ and for $t\in[N]$ we define the following random variables
\begin{align*}
I_t&=\textup{Coin selected at time $t$}\\
X_t&=\textup{Outcome of the toss at time $t$}\\
H_t&=(W,I_1,X_1,\ldots,I_{t},X_{t})
\end{align*}
where $H_t$ denotes the history till time $t$, the choice $I_t$ only depends on $H_{t-1}$.

For a bias vector $\boldp\in[0,1]^d$, along with the distribution on $W$ (which is independent of each coin), we let $P$ denote the probability distribution on the full transcript $H_N=(W,I_1,X_{1},\ldots,I_N,X_N)$ when the coins have bias $\boldp$. Furthermore, for $t\in[N]$ we let $P_t$ be distribution on $H_t=(W,I_1,X_1,\ldots,I_t,X_t)$ when the coins have bias $\boldp$, noting that $P_N=P$.

Similarly, for $\boldq\in[0,1]^d$ we let $Q$ denote the distribution on the full transcript induced when the coins have bias $\boldq$ and for $t\in[N]$, we let $Q_t$ be the distribution on $H_t=(W,I_1,X_1,\ldots,I_t,X_t)$ when the coins have bias $\boldq$.

Because $\calA$ is $(d+1,\varepsilon,\delta)$-list replicable, for each $\boldp\in[0,1]^d$ there exists a set $L_\boldp\subseteq\calY$ of size $|L_\boldp|\leq d+1$ such that each $\boldy\in L_\boldp$ it holds that $\lVert\boldp-\boldy\rVert_\infty\leq\varepsilon$ and 
\[\Pr(Y\in L_\boldp)\geq1-\delta\]

\begin{lemma}
Let $\boldp,\boldq\in[0,1]^d$. If $\mathbf{\frac{1}{4}}\leq\boldp\leq\mathbf{\frac{3}{4}}$ then $\EE_P\left[\left(\frac{dQ}{dP}\right)^2\right]\leq e^{\frac{16}{3}N\left\lVert\boldp-\boldq\right\rVert_{\infty}^2}$
\end{lemma}
\begin{proof}
Define $\Lambda_0\equiv1$ and, for $t\in[N]$, define
\[\Lambda_t=\frac{dQ_t}{dP_t}\]
and let $\pi_t(i\mid H_{t-1})$ denote the algorithm's probability of choosing coin $i\in[d]$ at time $t$, given the history $H_{t-1}$ up until that point. Conditioned on $H_{t-1}$, both $I_t$ and $\Lambda_{t-1}(H_{t-1})$ are fixed. Since the algorithm uses the same policy $\pi_t$, we have
\[\Pr\nolimits_P(I_t=i\mid H_{t-1})=\pi_t(i\mid H_{t-1})=\Pr\nolimits_Q(I_t=i\mid H_{t-1})\]
Because by definition, the algorithm's tosses can only depend on what it has seen so far. For any $t\in[N]$, we have
\begin{align*}
\Lambda_{t}(H_t)&=\Lambda_{t-1}(H_{t-1})\frac{\pi_t(I_t\mid H_{t-1})\cdot q_{I_t}^{X_t}(1-q_{I_t})^{1-X_t}}{\pi_t(I_t\mid H_{t-1})\cdot p_{I_t}^{X_t}(1-p_{I_t})^{1-X_t}}\\
&=\Lambda_{t-1}(H_{t-1})\frac{q_{I_t}^{X_t}(1-q_{I_t})^{1-X_t}}{p_{I_t}^{X_t}(1-p_{I_t})^{1-X_t}}\\
&=\begin{cases}
\Lambda_{t-1}(H_{t-1})\frac{q_{I_t}}{p_{I_t}}&\textup{if $X_t=1$}\\
\Lambda_{t-1}(H_{t-1})\frac{1-q_{I_t}}{1-p_{I_t}}&\textup{if $X_t=0$}
\end{cases}
\end{align*}
Observe the recurrence
\begin{align*}
&\EE_P[\Lambda_t^2(H_t)\mid H_{t-1}]\\
&\quad=p_{I_t}\cdot
\left(\Lambda_{t-1}(H_{t-1})\frac{q_{I_t}}{p_{I_t}}\right)^2+(1-p_{I_t})\cdot 
\left(\Lambda_{t-1}(H_{t-1})\frac{1-q_{I_t}}{1-p_{I_t}}\right)^2\\
&\quad=\Lambda_{t-1}^2(H_{t-1})\left(\frac{q_{I_t}^2}{p_{I_t}}+\frac{(1-q_{I_t})^2}{1-p_{I_t}}\right)\\
&\quad=\Lambda_{t-1}^2(H_{t-1})\left(1+\frac{(p_{I_t}-q_{I_t})^2}{p_{I_t}(1-p_{I_t})}\right)\\
&\quad\leq\Lambda_{t-1}^2(H_{t-1})e^{\frac{16}{3}\lVert\boldp-\boldq\rVert_\infty^2}
\end{align*}
The last three steps can be obtained by using a similar analysis to \Cref{lem:expectedsquaredderivatebernoullibound}. Finally, averaging over all possible histories and using the law of total expectation, we have
\begin{align*}
\EE_P[\Lambda_t^2(H_t)]  &=\EE_P[\EE_P[\Lambda_t^2(H_t)\mid H_{t-1}]]\\
                    &\leq\EE_P[\Lambda_{t-1}^2(H_{t-1})]e^{\frac{16}{3}\lVert\boldp-\boldq\rVert_\infty^2}
\end{align*}
A routine induction on $t$ shows that
\begin{align*}
\EE_P\left[\left(\frac{dQ}{dP}\right)^2\right]&=\EE_P[\Lambda_N^2(H_N)]\\
&\leq\EE_P[\Lambda_{0}^2(H_0)]\prod_{i=1}^{N}e^{\frac{16}{3}\lVert\boldp-\boldq\rVert_\infty^2}\\
&=1\cdot e^{\sum_{i=1}^{N}\frac{16}{3}\lVert\boldp-\boldq\rVert_\infty^2}\\
&=e^{\frac{16}{3}N\lVert\boldp-\boldq\rVert_\infty^2}\tag*{\qedhere}
\end{align*}
\end{proof}

The remaining analysis is identical to \Cref{subsec:lowerboundford1listreplicablealgorithms}. Use $\calA$ to color $\calR=\left[\frac{1}{2}-2\varepsilon,\frac{1}{2}+2\varepsilon\right]^d\subseteq\left[\frac{1}{4},\frac{3}{4}\right]^d$ with the most likely output, i.e., define the coloring $\chi\colon \calR\to\calY$ as
\[\chi(\boldp)=\argmax_{\boldy\in L_\boldp}\Pr\nolimits_P(Y=\boldy)\]
where ties are broken according to a fixed rule. Since $\chi$ is a valid Sperner/KKM coloring, the Neighborhood Sperner/KKM, \Cref{thm:neighborhoodspernerkkm}, applies. Similar to the analysis from \Cref{lem:lowerboundmainlemma}, we get
\[\frac{16}{3}Nr^2\geq \frac{dr}{10\varepsilon}-2\ln(2(d+1))\]
Picking $r=\frac{40\varepsilon}{d}\ln(2(d+1))$, for sufficiently large $d$, we again get
\[N\geq\frac{3}{12800}\cdot\frac{d^2}{\varepsilon^2\ln(2(d+1))}=\Omega\left(\frac{d^2}{\varepsilon^2\log d}\right)\]

We can again generalize this result to lists of size $k$ for arbitrary randomized algorithms.
\begin{theorem}[\Cref{thm:lowerboundmaintheoremklist} Restated]
Let $\varepsilon,\delta\in(0,1/8)$ and $d,k\in\mathbb{N}$. If $k\in[d+1,2^d]$ then every $(k,\varepsilon,\delta)$-list replicable algorithm for the $d$-$\CoinBiasEstimationProblem$ requires $N=\Omega\left(\frac{d^2}{\varepsilon^2\log k}\right)$ samples.
\end{theorem}
The argument is the same; the main bound that changes is
\[\Pr\nolimits_P(Y=\boldy_\boldp)\geq\frac{1-\delta}{|L_\boldp|}\geq\frac{7}{8k}\geq\frac{1}{2k}=:\beta\]
\Cref{lem:lowerboundmainlemma} now becomes
\[\frac{16}{3}Nr^2\geq \frac{dr}{10\varepsilon}-2\ln(2k)\]
If $k\leq\frac{1}{2}e^{\frac{d}{20}}$ then our choice $r:=\frac{40\varepsilon}{d}\ln(2k)\leq2\varepsilon$ holds and therefore
\[N\geq\frac{3}{12800}\cdot\frac{d^2}{\varepsilon^2\ln(2k)}=\Omega\left(\frac{d^2}{\varepsilon^2\log k}\right)\]
For $k>\frac{1}{2}e^{\frac{d}{20}}$, every $(k,\varepsilon,\delta)$-list replicable algorithm also solves the $(d,\varepsilon,\delta)$-$\CoinBiasEstimationProblem$. Using \Cref{lem:coinbiasestimationproblemsamplecomplexity} gives  
\[N=\Omega\left(\frac{d}{\varepsilon^2}\right)=\Omega\left(\frac{d^2}{\varepsilon^2\log k}\right)\]
The last bound was by using the fact that $k>\frac{1}{2}e^{\frac{d}{20}}$.

\end{document}